\PassOptionsToPackage{table, dvipsnames}{xcolor}
\documentclass[10pt,twocolumn,letterpaper]{article}

\usepackage[pagenumbers]{cvpr} 
\usepackage{soul}

\definecolor{cvprblue}{rgb}{0.21,0.49,0.74}
\usepackage[pagebackref,breaklinks,colorlinks,allcolors=cvprblue]{hyperref}

\usepackage{algorithm}
\usepackage[noend]{algpseudocode}
\usepackage{amsmath}
\usepackage{amssymb}
\usepackage[table]{xcolor}
\usepackage{multirow}
\usepackage{bm}

\definecolor{yellow}{rgb}{1, 1, 0.7}
\definecolor{orange}{rgb}{1, 0.85, 0.7}
\definecolor{red}{rgb}{1, 0.7, 0.7}

\DeclareMathOperator*{\argmin}{arg\,min}

\def\confName{CVPR}
\def\confYear{2025}

\title{
ZeroGS: Training 3D Gaussian Splatting from Unposed Images
}

\author{
    Yu Chen\textsuperscript{1} \quad 
    Rolandos Alexandros Potamias\textsuperscript{2} \quad
    Evangelos Ververas\textsuperscript{2} \quad \\
    Jifei Song\quad 
    Jiankang Deng\textsuperscript{2} \quad
    Gim Hee Lee\textsuperscript{1} \quad \\
    \textsuperscript{1}National University of Singapore \quad
    \textsuperscript{2}Imperial College of London \\ 
}

\begin{document}
\twocolumn[{%
\renewcommand\twocolumn[1][]{#1}%
\maketitle
\vspace{-25pt}
\begin{center}
    \captionsetup{type=figure}
    \vspace*{-6pt}
    \centering
    \includegraphics[width=0.98\linewidth]{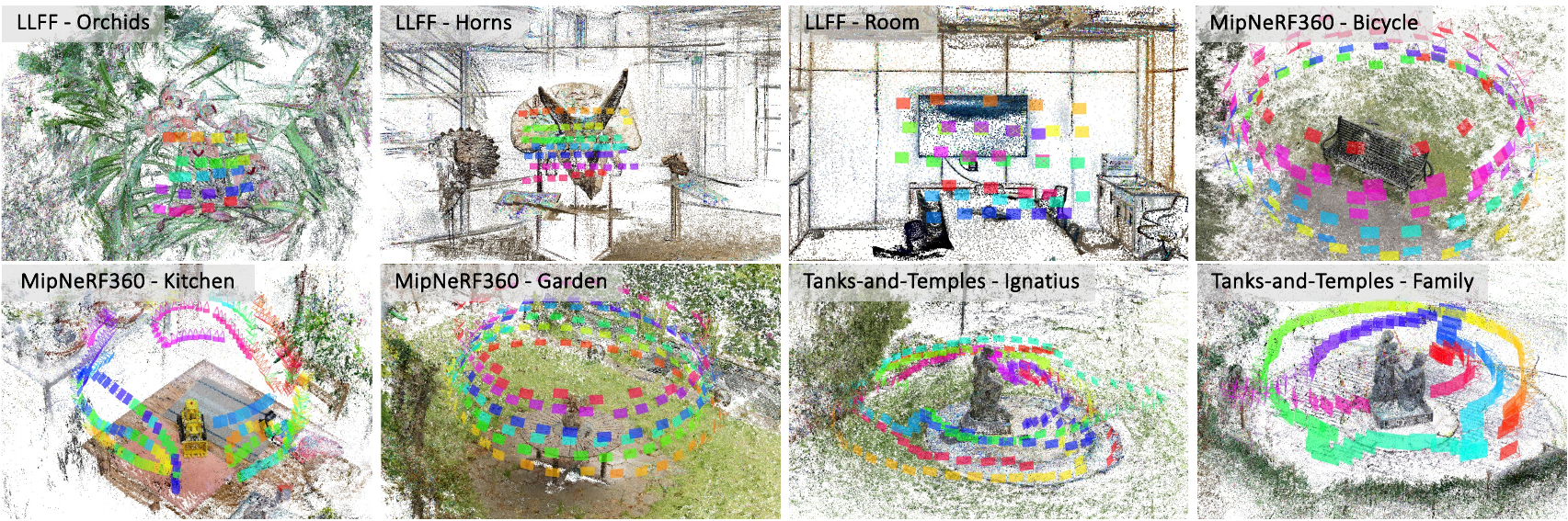}
    \vspace*{-6pt}
    \captionof{figure}{\textbf{Reconstruction results of ZeroGS}. Our method reconstructs scenes from hundreds of images without 
        COLMAP poses.
    }
    \vspace*{-5pt}
    \label{fig:teaser}
\end{center}%
}]

\begin{abstract}
Neural radiance fields (NeRF) and 3D Gaussian Splatting (3DGS) are popular techniques to reconstruct and render photo-realistic images. However, the pre-requisite of running Structure-from-Motion (SfM) to get camera poses limits their completeness. While previous methods 
can reconstruct from a few unposed images, they are not applicable when images are unordered or densely captured. In this work, we propose ZeroGS to train 3DGS from hundreds of unposed and unordered images. Our method leverages a pretrained foundation model as the neural scene representation. 
Since the accuracy of the predicted pointmaps does not suffice for accurate image registration and high-fidelity image rendering, we propose to mitigate the issue by initializing and finetuning the pretrained model from a seed image.
Images are then progressively registered and added to the training buffer, which is further used to train the model.
We also propose to refine the camera poses and pointmaps by minimizing a point-to-camera ray consistency loss across multiple views. Experiments on the LLFF dataset, the MipNeRF360 dataset, and the Tanks-and-Temples dataset show that our method recovers more accurate camera poses than state-of-the-art pose-free NeRF/3DGS methods, and even renders higher quality images than 3DGS with COLMAP poses. Our project page is available at \href{https://aibluefisher.github.io/ZeroGS/}{aibluefisher.github.io/ZeroGS}.
\end{abstract}    
\section{Introduction}
\label{sec:intro}

The renaissance of 3D reconstruction motivates many applications in recording real-world scenes and reconstructing them into a 3D digital world. Most travelers and tourists habitually take pictures and record videos at tourist attractions.
These pictures and videos are often uploaded onto websites or applications (such as Niantic Scaniverse\footnote{\url{https://nianticlabs.com/news/scaniverse4}} and PolyCam\footnote{\url{https://poly.cam/}}) and then reconstructed to 3D models. Behind these applications, neural radiance fields (NeRF)~\cite{DBLP:conf/eccv/MildenhallSTBRN20} and 3D Gaussian Splatting (3DGS)~\cite{DBLP:journals/tog/KerblKLD23} are increasingly becoming the most popular techniques used to reconstruct 3D scenes. While NeRF/3DGS can reconstruct photo-realistic scenes, they require accurate camera poses from Structure-from-Motion (SfM) tools, \eg, COLMAP~\cite{DBLP:conf/cvpr/SchonbergerF16}. NeRF/3DGS struggles 
on initialization with inaccurate camera poses and often produces blurry images. Existing methods try to optimize the inaccurate camera poses jointly with per-scene NeRF~\cite{DBLP:conf/iccv/LinM0L21,DBLP:journals/corr/abs-2204-05735,DBLP:conf/cvpr/ChenCWZ0S023,DBLP:journals/tmm/FuYLZ24}, or train a generalizable NeRF model~\cite{DBLP:conf/cvpr/ChenL23} or few-shot NeRF without ground-truth camera poses~\cite{DBLP:conf/cvpr/BianWLB23}. However, few works try to solve the same problem with 3DGS. While CF-3DGS~\cite{DBLP:conf/cvpr/Fu0LKKE24} can train 3DGS without relying on camera poses from COLMAP, it only works on short sequential images where camera poses do not change significantly between consecutive frames.

More recently, the 3D-vision foundation model DUSt3R~\cite{DBLP:conf/cvpr/Wang0CCR24} has motivated methods to decouple camera poses from the training process of generalizable 3DGS~\cite{DBLP:journals/corr/abs-2408-13912,ye2024noposplat}. DUSt3R takes an image pair as input and outputs pairwise pointmaps in the coordinate frame of the reference image. DUSt3R is trained with massive real-world data containing accurate camera poses and 3D points and generalizes very well to unseen image pairs.
However, these generalizable 3DGS methods~\cite{DBLP:journals/corr/abs-2408-13912,ye2024noposplat} can only handle the few-shot setting of an image pair since they heavily rely on DUSt3R. InstantSplat~\cite{fan2024instantsplat} leverages the pretrained DUSt3R as a offline tool. Given pairwise images from the same scene, InstantSplat first computes the pairwise pointmaps with DUSt3R, and then aligns the dense pointmaps by jointly optimizing the camera poses and dense points into a global coordinate frame. Subsequently, InstantSplat trains 3DGS with the aligned dense points from DUSt3R. However, the huge GPU memory requirement in optimizing dense pointmaps limits InstantSplat to scenes with 
few images.

In this work, we propose \textbf{ZeroGS} to train 3DGS without relying on COLMAP camera poses. Unlike CF-3DGS and InstantSplat 
which can only be used for short image sequences or a few images, our method can reconstruct scenes from hundreds of unordered images (\cf Fig.~\ref{fig:teaser}). Specifically, we leverage a pretrained 3D foundation model as our neural scene representation. In addition to predicting pointmaps, we extend the model to predict the properties of 3D Gaussian primitives. The pretrained model has learned coarse scene geometry priors, making it much easier to jointly optimize the model and camera parameters from scratch. After defining our neural scene representation, we adopt an incremental training pipeline to finetune our model. 
We first register images by computing coarse camera poses with RANSAC and PnP using the predicted pointmaps in the global coordinate frame. The coarse camera poses are then refined by a point-to-camera ray consistency loss, and our model is further finetuned on the newly registered images. We repeat the process until all images are registered. In this way, our incremental training pipeline is similar to the classical incremental SfM method~\cite{DBLP:conf/cvpr/SchonbergerF16} but differs as follows:
\begin{itemize}
  \item \textbf{Seed Initialization}. Incremental SfM initializes from an image pair, where the camera poses of the image pair are fixed after initialization to 
 fix the gauge freedom. However, our method initializes from a pretrained model and only one seed image.
  \item \textbf{Image Registration}. Instead of registering images 
  individually in an incremental manner, our method registers a batch of images each time. The registered images are further utilized to finetune the neural model.
  \item \textbf{Objective Function}. We finetune our model using a rendering loss but scenes are optimized by the reprojection error in SfM. 
  \item \textbf{Scene Sparsity}. Our model predicts dense scene geometries, while SfM outputs sparse scene structures.
\end{itemize}

Our incremental training pipeline also shares some similarities with a recent learning-based SfM method ACE0~\cite{DBLP:journals/corr/abs-2404-14351}, with several key differences: 1) ACE0 only predicts sparse pointmaps, while our method predicts dense pointmaps and 3D Gaussian primitives. 2) ACE0 uses 2D CNN and MLP as the neural scene representation, while we use transformers as the scene representation. 3) The training batch of ACE0 is composed of pixels from multiple views, while our method takes as input image pairs in the training batch. Moreover, finetuning a pretrained foundation model such as DUSt3R is not easy. This is because these 3D foundation models are supervised by ground-truth 3D points that are difficult to obtain in unseen scenes.

We evaluated our method on the LLFF~\cite{DBLP:journals/tog/MildenhallSCKRN19} dataset, the MipNeRF360~\cite{DBLP:conf/cvpr/BarronMVSH22} dataset, and the Tanks-and-Temples dataset~\cite{Knapitsch2017} and compared them to state-of-the-art pose-free NeRF/3DGS methods. The experimental results show that our method is the best among these methods and our image rendering quality is even better than training NeRF/3DGS from the COLMAP poses.

\section{Related Work}
\label{sec:related_work}

\paragraph{Neural Radiance Fields.}
Neural radiance fields~\cite{DBLP:conf/eccv/MildenhallSTBRN20} enable rendering from novel viewpoints with encoded frequency features~\cite{DBLP:conf/nips/TancikSMFRSRBN20}. Many follow-up works try to improve the rendering and training efficiency~\cite{DBLP:conf/nips/LiuGLCT20,DBLP:conf/iccv/YuLT0NK21,DBLP:conf/cvpr/Fridovich-KeilY22,DBLP:conf/eccv/ChenXGYS22} by encoding scenes into sparse voxels, multi-resolution hash tables~\cite{DBLP:journals/tog/MullerESK22}, or three orthogonal axes and planes. Another branch of NeRF methods focuses on generalizable NeRF~\cite{DBLP:conf/cvpr/WangWGSZBMSF21,DBLP:conf/iccv/ChenXZZXY021,DBLP:conf/cvpr/JohariLF22,DBLP:conf/cvpr/SuhailESM22,DBLP:conf/cvpr/LiuPLWWTZW22}, and alleviating the aliasing issue by approximating the cone sampling into the scale-aware integrated positional encodings~\cite{DBLP:conf/iccv/BarronMTHMS21} for vanilla NeRF or hexagonal sampling~\cite{DBLP:conf/iccv/BarronMVSH23} for Instant-NGP~\cite{DBLP:journals/tog/MullerESK22}.
Works are also done in registering multiple blocks of NeRF using traditional optimization method~\cite{DBLP:conf/icra/GoliRSGT23} or pretrain a generalizable geometry-aware transformer~\cite{DBLP:conf/iccv/ChenL23} from 3D data. Despite the limitation of training on small-scale scenes, the divide-and-conquer strategy is adopted to handle city-scale scenes~\cite{DBLP:conf/cvpr/TancikCYPMSBK22,DBLP:conf/cvpr/TurkiRS22,DBLP:conf/cvpr/RematasLSBTFF22,DBLP:conf/iclr/Mi023,DBLP:conf/cvpr/XuXPPZT0L23}.

To remove SfM poses from the training pipeline, NeRF$--$~\cite{DBLP:journals/corr/abs-2102-07064} jointly optimizing the network of NeRF and camera pose embeddings, SiNeRF~\cite{DBLP:journals/corr/abs-2210-04553} adopts a 
SIREN-MLP~\cite{DBLP:conf/nips/SitzmannMBLW20} and a mixed region sampling strategy to circumvent the sub-optimality issue in NeRF$--$. BARF~\cite{DBLP:conf/iccv/LinM0L21} proposes joint training of NeRF with imperfect camera poses from coarse-to-fine, where high-frequencies are progressively activated during training to alleviate the gradient inconsistency issue. GARF~\cite{DBLP:journals/corr/abs-2204-05735} extends BARF with the Gaussian activation, enabling training a positional-embedding less coordinate network. RM-NeRF~\cite{DBLP:journals/corr/abs-2210-04233} jointly trains a GNN-based motion averaging network~\cite{DBLP:conf/cvpr/Govindu04,DBLP:conf/eccv/PurkaitCR20} and Mip-NeRF~\cite{DBLP:conf/iccv/BarronMTHMS21} to solve the camera pose refinement issue in multi-scale scenes. However, all the above methods can only handle simple scenes (\eg, forward-facing cameras only) or require accurate pose priors to converge.

\paragraph{3D Gaussian Splatting.}
Different from NeRF which uses volume rendering to infer the scene occupancy, 3D Gaussian Splatting~\cite{DBLP:journals/tog/KerblKLD23} (3DGS) initializes 3D Gaussians from a sparse point cloud and renders scenes by differentiable rasterization, and can achieve real-time rendering performance. However, 3DGS can face difficulty in identifying more Gaussians when initialized from textureless areas.
To encourage the learning of a better scene geometry,
Scaffold-GS~\cite{DBLP:conf/cvpr/0005YXX0L024} initializes a sparse voxel grid from the initial point cloud and encodes the features of 
3D Gaussians into corresponding feature vectors. The introduction of the sparse voxel reduces the Gaussian densities by avoiding unnecessary densification on the scene surface. SAGS~\cite{DBLP:journals/corr/abs-2404-19149} implicitly encodes the scene structure into a GNN. Other works also try to learn 2D Gaussians~\cite{DBLP:conf/siggraph/HuangYC0G24} to fit the scene surface~\cite{DBLP:conf/cvpr/GuedonL24} more accurately. Similar to NeRF, 3DGS faces the aliasing issue caused by the fixed window Gaussian kernel during rasterization. The same issue is handled in Mip-Splatting~\cite{DBLP:conf/cvpr/YuCHS024} and many latter works~\cite{DBLP:conf/cvpr/YanLCL24,DBLP:journals/corr/abs-2403-11056,DBLP:journals/corr/abs-2403-19615}. VastGaussian~\cite{DBLP:conf/cvpr/LinLTLLLLWXYY24} and its follow-ups~\cite{DBLP:journals/tog/KerblMKWLD24,DBLP:journals/corr/abs-2404-01133,DBLP:journals/corr/abs-2405-13943} focus on developing distributed training methods to reconstruct the large-scale scenes. 

Although 3DGS can render higher-fidelity images, it relies on accurate camera poses. Unlike NeRF where many works have been proposed to solve the inaccuracy of camera poses, the same issue has not been widely tackled in 3DGS. The most relevant work to ours is CF-3DGS~\cite{DBLP:conf/cvpr/Fu0LKKE24} and InstantSplat~\cite{fan2024instantsplat}. However, CF-3DGS requires depth estimation to initialize the 3D Gaussians, and it can only optimize camera poses and 3DGS between short consecutive image sequences. 
CF-3DGS is highly susceptible to failure when camera poses change significantly or images are unordered.
InstantSplat~\cite{fan2024instantsplat} leverages an existing pretrained 3D foundation model DUSt3R~\cite{DBLP:conf/cvpr/Wang0CCR24} to regress dense pointmaps between image pairs, followed by obtaining the camera poses by aligning the pointmaps into a global coordinate frame. However, aligning dense pointmaps is time- and memory-consuming. As a result, InstantSplat can only handle a very few images. Other related work includes Splatt3R~\cite{DBLP:journals/corr/abs-2408-13912} which extends DUSt3R to predict 3D Gaussians without posed images. Nonetheless, it works only on image pairs since DUSt3R produces pointmaps in the reference frame of the first image instead of a global coordinate frame.
Our work also leverages a pretrained DUSt3R-based network, \ie Spann3R~\cite{DBLP:journals/corr/abs-2408-16061}. However, unlike InstantSplat and Splatt3R, our method regresses points in the global coordinate frame of a seed image without aligning the dense pointmaps, followed by incrementally registering images and finetuning the network. Our method can handle hundreds of images and run on a 24GB consumer-level GPU.

\section{Method}
\label{sec:method}

\begin{figure*}[htbp]
    \centering
    \includegraphics[width=1.0\linewidth]{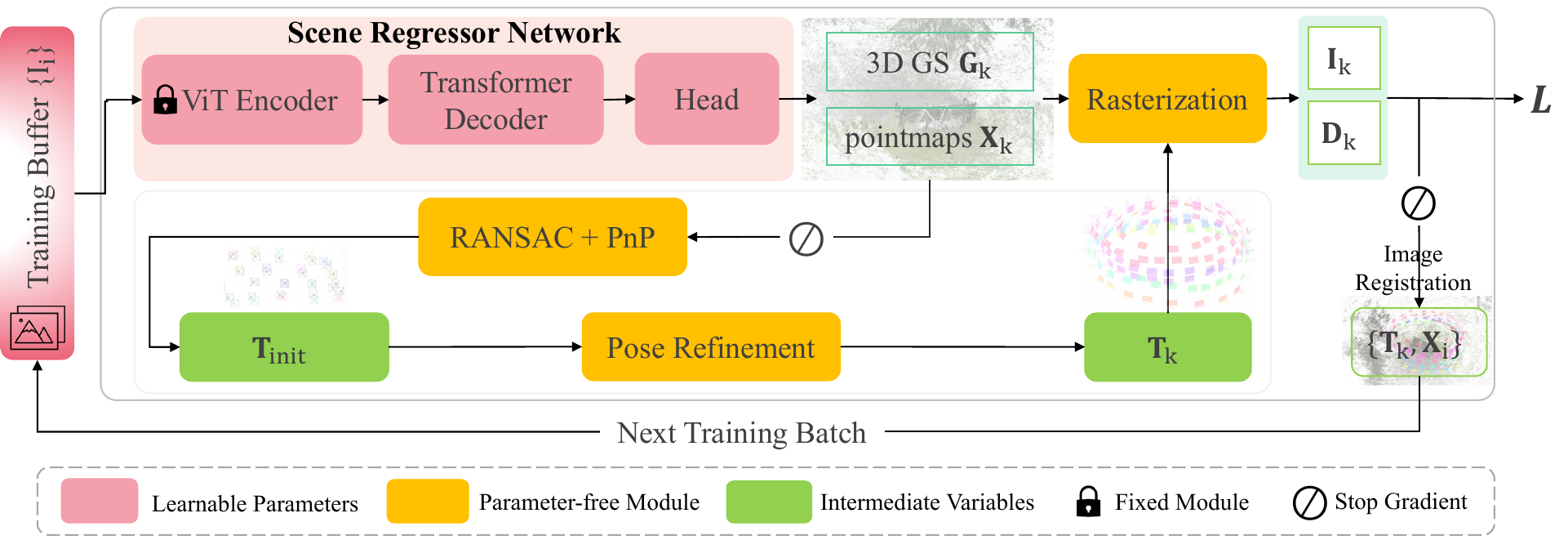}
    \vspace{-5mm}
    \caption{\textbf{The training pipeline of our Pose-Free 3D Gaussian Splatting}.
         Our method follows the classical incremental SfM reconstruction pipeline with the key difference that the input is no longer an image but a pair of images in a progressively updated training buffer. The scene regressor network is trained as follows:
         1) Use Spann3R~\cite{DBLP:journals/corr/abs-2408-16061} as the scene regressor network to predict 3D Gaussians $\mathbf{G}_k$ and pointmaps $\mathbf{X}_k$ from a pair of images.
         2) Leverage RANSAC and a PnP solver to obtain the initial camera poses based on direct 2D-3D correspondences. 
         3) Refine the coarse camera poses by minimizing the point-to-ray consistency loss between 3D tracks and camera centers. 
         4) Rasterize the 3D Gaussians with the refined camera poses to render images. An RGB loss is adopted for back-propagating gradients. 
         5) After each training epoch, we update the training buffer by registering more images.
    }
    \label{fig:zero_gs_network}
    \vspace{-3mm}
\end{figure*}

In this section, we first give the preliminaries of our scene regressor network followed by introducing our training pipeline (see Fig.~\ref{fig:zero_gs_network}) for incrementally reconstructing a scene.
We first use Spann3R as the scene regressor network to predict 3D Gaussians $\mathbf{G}_k$ and 
pointmaps $\mathbf{X}_k$ from an image pair.  We then use RANSAC and a PnP solver to obtain the initial camera poses based on 2D-3D correspondences. Furthermore, we refine the coarse camera poses by minimizing a point-to-camera ray consistency loss between 3D points and camera centers. Subsequently, we rasterize the 3D Gaussians with the
refined camera poses to render images. An RGB loss is adopted for back-propagating gradients. At the end of each training epoch, we update our training buffer by registering new images with RANSAC and a PnP solver. This process is repeated until all images are registered or no more images can be registered.

\subsection{Preliminaries}

Our network architecture is based on DUSt3R~\cite{DBLP:conf/cvpr/Wang0CCR24} and 
Spann3R~\cite{DBLP:journals/corr/abs-2408-16061}. Given an image pair $( \mathbf{I}_i, \mathbf{I}_j )$, DUSt3R predicts the corresponding pointmaps $( \mathbf{X}^{i,i}, \mathbf{X}^{j,i} )$ for each image, where $\mathbf{X}^{i,j}$ denotes the pointmap $\mathbf{X}_j$ expressed in camera $i$'s coordinate frame. 

\vspace{1mm}
\noindent \textbf{DUSt3R} uses a ViT~\cite{DBLP:conf/iclr/DosovitskiyB0WZ21} as a shared encoder for both images and two transformer decoders for the reference image $i$ and the target image $j$, respectively. The two decoders denoted as \textit{reference decoder} $\mathcal{D}_{\text{ref}}$ and \textit{target decoder} $\mathcal{D}_{\text{tgt}}$ consist of two projection heads $\mathcal{H}_{\text{ref}},\mathcal{H}_{\text{tgt}}$ that map the decoder features into pointmaps:
\begin{align}
    & \mathbf{f}_i^{\text{e}},\mathbf{f}_j^{\text{e}} = \mathcal{V}(\mathbf{I}_i, \mathbf{I}_j), 
    \mathbf{f}_i^{\text{d}} = \mathcal{D}_{\text{ref}}(\mathbf{f}_i^{\text{e}}, \mathbf{f}_j^{\text{e}}),
      \mathbf{f}_j^{\text{d}} = \mathcal{D}_{\text{tgt}}(\mathbf{f}_j^{\text{e}}, \mathbf{f}_i^{\text{e}}), \\ \nonumber
    & \mathbf{X}^{i,i},\ \mathbf{X}^{j,i} = \mathcal{H}_{\text{ref}}(\mathbf{f}_i^{\text{d}}),\ \mathcal{H}_{\text{tgt}}(\mathbf{f}_j^{\text{d}}).
\end{align}
DUSt3R reconstructs image pairs in a local coordinate frame. When handling more than two images, DUSt3R uses a post-processing step to align the pairwise dense pointmaps to a global coordinate frame, which is time-consuming and can exceed the GPU memory limitation.

\vspace{1mm}
\noindent \textbf{Spann3R} proposes a feature fusion mechanism to predict pointmaps $(\mathbf{X}^{i,g},\ \mathbf{X}^{j,g})$ in a global coordinate frame. It computes a fused feature $\mathbf{f}^{G}_t$ in the $t$-th training epoch from a spatial feature memory. The reference decoder inputs the fused feature for reconstruction and the target decoder produces features for querying the memory. 
%
%
Furthermore, Spann3R uses two additional projection heads to compute the key and query feature for reconstructing the next image pairs, and a memory encoder $\mathcal{V}_{\text{mem}}$ which encodes the pointmaps from the reference decoder:
\begin{align}
    & \mathbf{f}^{Q}_j = \mathcal{H}_{\text{tgt}}^{\text{Q}} (\mathbf{f}_j^d),\ 
      \mathbf{f}^{K}_i = \mathcal{H}_{\text{ref}}^{\text{K}} (\mathbf{f}_i^{\text{d}}),\
    \mathbf{f}^{V}_i = \mathcal{V}_{\text{mem}} (\mathbf{X}^{i,g}).
\end{align}
Although Spann3R can reconstruct out-of-distributed scenes, it reconstructs images individually and is limited to very short frames due to GPU memory limitation. Moreover, the 3D points predicted by Spann3R in these scenes lack accuracy. We refer readers to~\cite{DBLP:conf/cvpr/Wang0CCR24,DBLP:journals/corr/abs-2408-16061} for more details.

\subsection{Neural Scene Representation}
In 3DGS, scenes are explicitly represented by a set of 3D Gaussian primitives~\cite{DBLP:journals/tog/KerblKLD23} or implicitly represented by neural networks~\cite{DBLP:conf/cvpr/0005YXX0L024,octreegs}. In this work, we use Spann3R~\cite{DBLP:journals/corr/abs-2408-16061} as neural scene representation and extend it to also predict Gaussian primitives. We refer to our neural scene representation as the \emph{scene regressor network} $f_{\text{SCR}}$, which analogs to the scene regressor in pose regression or localization networks~\cite{DBLP:conf/cvpr/BrachmannCP23,DBLP:journals/corr/abs-2404-14351}. Unlike the sparse scene regressor that takes an image as input, our scene regressor takes an image pair as input and predicts dense pointmaps $\mathbf{X}_i$ and per-pixel 3D Gaussians $\mathbf{G}_i$ in a global coordinate frame:
\begin{equation}
    \left( \mathbf{X}_i, \mathbf{G}_i;\ \mathbf{X}_j, \mathbf{G}_j \right) = f_{\text{SCR}}(\mathbf{I}_i, \mathbf{I}_j),
\end{equation}
where $\mathbf{I}_i$ is the reference and  $\mathbf{I}_j$ is the target image.

By rasterizing the set of 3D Gaussian primitives $\mathcal{G} = \{\mathbf{G}_i\}$, we back-propagate gradients to the model using an RGB loss. More specifically, a 3D Gaussian primitive is composed of the opacity $o$, the mean $\mathbf{u}$, the covariance $\mathbf{\Sigma}$, and the coefficients of the spherical harmonics $\mathbf{SH}$. The covariance is decomposed into a rotation matrix $\mathbf{R}$ and a scaling matrix 
$\mathbf{S}$ to ensure the positive semi-definiteness: $\mathbf{\Sigma}_i = \mathbf{R} \mathbf{S} \mathbf{S}^{\top} \mathbf{R}^{\top}$. 
In addition, instead directly predicting the mean $\mathbf{u}$ for each 3D Gaussian, we predict an offset 
$\Delta \mathbf{X}$ and apply it to the pointmaps to obtain the mean $\mathbf{u} = \mathbf{X} + \Delta \mathbf{X}$.
To render the color for a pixel $\mathbf{p}$, the 3D Gaussians are projected into the image space for alpha blending:
\begin{equation}
 \mathbf{C} = \sum_{i} \mathbf{c}_i \alpha_i \prod_{j=1}^{i-1} (1 - \alpha_j),
 \label{eq:alpha_composition}
\end{equation}
where $\alpha_i$ is the rendering opacity and is computed by $\alpha = o \cdot \mathbf{G}^{\text{proj}} (\mathbf{p})$, $\mathbf{c}_i$ is the per-pixel color that computed from the spherical harmonics $\mathbf{SH}$. In practice, Eq.~\eqref{eq:alpha_composition} is computed using a differentiable rasterizer~\cite{DBLP:journals/tog/KerblKLD23}:
\begin{equation}
    \hat{\mathbf{I}} = \mathcal{R} (\mathbf{T}, \mathbf{K}; o,\mathbf{R},\mathbf{S},\mathbf{SH}) = \mathcal{R}(\mathbf{T}, \mathbf{K}; \{\mathbf{G}_i\}),
\end{equation}
where $\mathbf{T}$ is the camera extrinsics and $\mathbf{K}$ is the intrinsics.

\subsection{Incremental Reconstruction}
\label{subsec:incremental_recon}

We incrementally reconstruct each scene with the scene regressor as the neural representation.
We emphasize that finetuning a pre-trained model such as DUSt3R or Spann3R on unseen scenes is challenging. This is because 
the existing DUSt3R-based model is trained with ground-truth 3D points. However, obtaining ground-truth 3D points can be expensive and impossible in most scenes.

\begin{figure*}[htbp]
    \centering
    \includegraphics[width=1.0\linewidth]{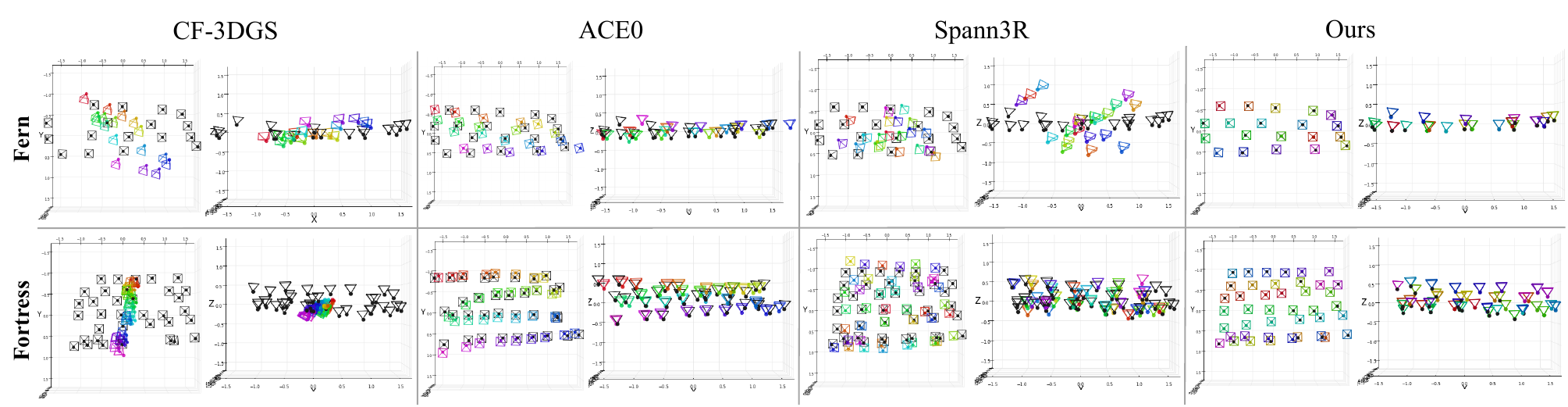}
    \vspace{-5mm}
    \caption{\textbf{Visualization of camera poses accuracy} on the LLFF dataset (Zoom in for best view).
     Black: pseudo-ground-truth camera poses obtained from COLMAP~\cite{DBLP:conf/cvpr/SchonbergerF16}. Colored: predicted camera poses.
    }
    \label{fig:llff_pose_cmp}
    \vspace{-5mm}
\end{figure*}

\subsubsection{Seed Initialization}
\label{subsubsec:seed_init}
Given a set of unordered images $\mathcal{I}=\{\mathbf{I}_i\}$, we first select a \textit{seed image} for initialization. This is different from incremental 
SfM, which requires a \textit{seed image pair} for two-view reconstruction. The seed image pair must have sufficient matching inliers and a wide baseline. This criterion guarantees the initial pair overlaps with as many other images as possible for later registration. To achieve a similar goal, we use NetVLAD~\cite{DBLP:conf/cvpr/ArandjelovicGTP16} to compute a global descriptor for each image, and then we compute the similarity score between each image pair. We further build a similarity graph $\mathcal{G}_{\text{sim}}$, where the node represents an image, the edge represents the image pair, and the edge weight represents the similarity score. An edge is discarded if its weight is less than a threshold $s_{\text{sim}}$. We then select the node that has the maximum degree as the seed image $\mathbf{I}_{\text{seed}}$. Intuitively, a node with a maximum degree means it has the most adjacent images, which is beneficial for the batched registration of images.

After selecting the seed image, we finetune the scene regressor in a self-supervised manner. Specifically, we set the seed image as the reference frame, and the camera pose of the seed image, $\mathbf{T}_{\text{seed}}$, is set to an identity matrix. We then compute a RGB loss:
\vspace{-1mm}
\begin{small}
\begin{align}
    & \mathcal{L}_{\text{rgb}}
    = \sum \| \mathbf{I} - \hat{\mathbf{I}} \|_1
    = \sum \| \mathbf{I} - \mathcal{R}(\mathbf{T}_{\text{seed}}, \mathbf{K}, \{\mathbf{G}_i\}) \|_1.
    \label{eq:rgb_loss}
\end{align}
\end{small}
Note that, during initialization, the seed image serves as both the reference and target image to the scene regressor:
$\left( \mathbf{X}_i, \mathbf{G}_i;\ \mathbf{X}_j, \mathbf{G}_j \right) = f_{\text{SCR}}(\mathbf{I}_{\text{seed}}, \mathbf{I}_{\text{seed}})$, and 
the camera pose is fixed as an identity matrix.

\subsubsection{Image Registration}
After seed initialization, we incrementally register a batch of images $\mathcal{I}_{\text{buf}}=\{\mathbf{I}_{k}\}$ in a training epoch. We add a batch of newly registered images into the training buffer and train the scene regressor. Upon training convergence, we expand the training buffer by selecting a new batch of images. This process is repeated until all images are registered.

\vspace{-3mm}
\paragraph{Coarse Camera Pose Estimation.}
Given a registered reference image $\mathbf{I}_i$ and a to-be-registered target image $\mathbf{I}_k$, we pass them to the scene regressor and obtain the 3D points $\{\mathbf{X}_k\}$ in a global coordinate frame. Since we have the coordinates $\{\mathbf{u}_k\}$ of each image pixel and their corresponding 3D coordinates $\{\mathbf{X}_k\}$, we can easily find the 2D-3D correspondences $\{(\mathbf{u}_k, \mathbf{X}_k)\}$. We then use RANSAC and a PnP solver to obtain a coarse camera pose:
\begin{small}
\begin{equation}
    \mathbf{T}_k^{\text{coarse}}, S_k = \text{PnP} (\mathbf{K}, \{(\mathbf{u}_k, \mathbf{X}_k)\}),
    \label{eq:coarse_camera_pose}
\end{equation}
\end{small}
where $S_k$ is the number of inliers and 
$\mathbf{X}_k=f_{\text{SCR}}(\mathbf{I}_{\text{ref}}, \mathbf{I}_k)$. $\mathbf{I}_{\text{ref}}$ is the reference image and $\mathbf{I}_k$ is the target image we want to register. We add the target image $\mathbf{I}_k$ into the training buffer only when the inlier number is larger than the inlier threshold $s_{\text{inlier}}$. After initialization, the seed image is selected as the reference image. In the following training batches, we select the reference image from the registered images which connects to most of the unregistered images.

\begin{table*}[htbp]
    \centering

\resizebox{0.95\textwidth}{!}{
    \begin{tabular}{l | c c  c c | c c | c c  c c | c c}
        \toprule
        
        \multirow{2}{*}{\textbf{Scenes}} &
        \multicolumn{2}{c}{\textbf{BARF}~\cite{DBLP:conf/iccv/LinM0L21}} &
        \multicolumn{2}{c}{\textbf{DBARF}~\cite{DBLP:conf/cvpr/ChenL23}} $\ \ |$ & 
        \multicolumn{2}{c}{\textbf{ACE0}~\cite{DBLP:journals/corr/abs-2404-14351}} $\ \ |$&
        \multicolumn{2}{c}{\textbf{CF-3DGS}~\cite{DBLP:conf/cvpr/Fu0LKKE24}} &
        \multicolumn{2}{c}{\textbf{Spann3R}~\cite{DBLP:journals/corr/abs-2408-16061}} $\ \ |$ &
        \multicolumn{2}{c}{\textbf{Ours}} \\
        
        \cmidrule(r){2-3} \cmidrule(r){4-5} \cmidrule(r){6-7} \cmidrule(r){8-9} \cmidrule(r){10-11} \cmidrule(r){12-13} 
       
        & \multirow{1}{*}{$\Delta \mathbf{R}$}
        & \multirow{1}{*}{$\Delta \mathbf{t}$} 
        
        & \multirow{1}{*}{$\Delta \mathbf{R}$}
        & \multirow{1}{*}{$\Delta \mathbf{t}$} 

        & \multirow{1}{*}{$\Delta \mathbf{R}$}
        & \multirow{1}{*}{$\Delta \mathbf{t}$} 
        
        & \multirow{1}{*}{$\Delta \mathbf{R}$}
        & \multirow{1}{*}{$\Delta \mathbf{t}$} 
       
        & \multirow{1}{*}{$\Delta \mathbf{R}$}
        & \multirow{1}{*}{$\Delta \mathbf{t}$} 
    
        & \multirow{1}{*}{$\Delta \mathbf{R}$}
        & \multirow{1}{*}{$\Delta \mathbf{t}$}  \\
       
        \midrule
       
        Fern
        & \cellcolor{red}0.19 & \cellcolor{orange}0.192  
        & \cellcolor{yellow}0.89 & 0.341 
        & 11.87 & \cellcolor{yellow}0.284  
        & 2.81 & 9.254
        & 39.03 & 0.767
        & \cellcolor{orange}0.26 & \cellcolor{red}0.005 \\
       
    
        Flower
        & \cellcolor{orange}0.25 & 0.224  
        & 1.39 & 0.318 
        & 10.32 & \cellcolor{orange}0.103  
        & \cellcolor{red}0.24 & 2.586 
        & 11.91 & \cellcolor{yellow}0.285 
        & \cellcolor{yellow}0.52 & \cellcolor{red}0.011 \\
       
        
        Fortress
        & \cellcolor{orange}0.48 & 0.364  
        & \cellcolor{yellow}0.59 & \cellcolor{yellow}0.229 
        & 75.07 & 0.603  
        & 1.28 & 8.592 
        & 08.31 & \cellcolor{orange}0.152 
        & \cellcolor{red}0.04 & \cellcolor{red}0.002 \\
    
       
        Horns
        & \cellcolor{orange}0.30 & \cellcolor{orange}0.222  
        & \cellcolor{yellow}0.82 & 0.292 
        & 07.61 & \cellcolor{yellow}0.233  
        & 1.15 & 2.371 
        & 06.98 & 0.349 
        &  \cellcolor{red}0.03 & \cellcolor{red}0.001 \\
    
    
        Leaves
        & \cellcolor{yellow}1.27 & \cellcolor{yellow}0.249  
        & 4.63 & 0.855 
        & 10.87 & \cellcolor{orange}0.136  
        & \cellcolor{orange}0.33 & 7.350 
        & 44.09 & 0.801 
        &  \cellcolor{red}0.22 & \cellcolor{red}0.006 \\
    
    
        Orchids
        & \cellcolor{orange}0.63 & 0.404  
        & \cellcolor{yellow}1.16 & 0.573 
        & 06.24 & \cellcolor{orange}0.168  
        & 1.45 & 2.772 
        & 09.77 & \cellcolor{yellow}0.256
        & \cellcolor{red}0.24 & \cellcolor{red}0.006 \\
    
    
        Room
        & \cellcolor{orange}0.32 & \cellcolor{orange}0.270  
        & \cellcolor{yellow}0.53 & \cellcolor{yellow}0.360 
        & 13.19 & 0.424  
        & 1.36 & 3.336 
        & 07.48 & 0.513
        &  \cellcolor{red}0.03 & \cellcolor{red}0.001 \\
    
    
        Trex
        & \cellcolor{orange}0.14 & 0.720  
        & \cellcolor{yellow}1.06 & \cellcolor{yellow}0.463 
        & 11.93 & \cellcolor{orange}0.373  
        & 1.56 & 4.431 
        & 32.39 & 0.758
        &  \cellcolor{red}0.03 & \cellcolor{red}0.010 \\
       
        \bottomrule
    \end{tabular}
}
    \vspace{-3mm}
    \caption{\textbf{Quantitative results of camera pose accuracy on LLFF dataset}.
     The \colorbox{red}{red}, \colorbox{orange}{orange} and \colorbox{yellow}{yellow} colors respectively 
     denote the best, the second best, and the third best results. The unit for rotation error is degree.
    }
    \label{table:quantitative_llff_camera_pose_acc}
    \vspace{-5mm}
\end{table*}

\vspace{-3mm}
\paragraph{Camera Pose Refinement.}
The camera poses of newly registered images can be inaccurate since the scene regressor has not seen these images. While ACE0~\cite{DBLP:journals/corr/abs-2404-14351} uses a MLP pose refiner to alleviate this issue during training, we experimentally found that does not improve the pose accuracy with our transformer-based scene regressor. This is because ACE0 uses MLP as the scene coordinate decoder, and each pixel is individually mapped onto the 3D space. ACE0 thus enables network training by mixing millions of pixels from different views in a training batch, and the multiple-view constraint helps constrain the network training. However, since we use a transformer-based decoder and due to the GPU memory limitation, we can use only limited views (we use a batch size 1 in practice on a 24GB consumer-level GPU) in each training batch, which can easily lead to the divergence of network training.

To solve the aforementioned issue, we propose to further refine the coarse camera poses by minimizing a point-to-camera ray consistency loss below:
\begin{small}
\begin{equation}
   \argmin_{\mathbf{X}_i, \mathbf{C}_k} \sum_{i,k} \rho( \| d_{i,k} \cdot \bm{\nu}_{i,k} - (\mathbf{X}_i - \mathbf{C}_k) \|_2),
   \label{eq:point_to_camera_ray_loss}
\end{equation}
\end{small}
where $\mathbf{C}_k$ is the camera center for image $\mathbf{I}_k$, $d_{i,k}$ is the scaling factor between a 3D point $\mathbf{X}_i$ and the camera center $\mathbf{C}_k$, $\bm{\nu}_{i,k}$ is the ray direction between $\mathbf{X}_i$ and $\mathbf{C}_k$. During optimization, we fix the camera pose of the seed image for the gauge ambiguity and fix the scaling factor between the seed image and its most similar adjacent image for the scale ambiguity. Moreover, for a new training epoch, we fix the camera poses registered in the previous epoch and only optimize the camera poses registered in the current training epoch to improve the optimization efficiency.

\subsubsection{Finalizing Neural Scene Reconstruction}
\label{subsubsec:finalize_neural_scene}
We propose a two-stage strategy to improve the final reconstruction quality when all images have been registered or no more images can be added to the training buffer. The first stage is to optimize all camera poses using Eq.~\eqref{eq:point_to_camera_ray_loss}. This is because we incrementally register images and errors accumulate during training. In this stage, we only fix the camera pose of the seed image, and camera poses obtained from all previous training epochs are used as initial values for optimization. Since the initial values are accurate enough, the final optimization converges very fast. To further improve the image rendering quality, we proposed to refine the scene details using explicit 3D Gaussian primitives~\cite{DBLP:journals/tog/KerblKLD23} in a second stage. This is because we only used fixed low-resolution images during the training of our scene regressor due to GPU memory limitation. The scene regressor can therefore only represent the coarse scene geometry. In the second stage, we use the same strategy as in~\cite{DBLP:journals/tog/KerblKLD23} for 3D Gaussian densification and pruning during refinement.

\section{Experiments}
\label{sec:experiments}

\paragraph{Evaluation Datasets.}
We evaluate our method on the LLFF dataset~\cite{DBLP:journals/tog/MildenhallSCKRN19}, the Mip-NeRF360 dataset~\cite{DBLP:conf/cvpr/BarronMVSH22}, and the Tanks-and-Temples dataset~\cite{Knapitsch2017}. The LLFF dataset contains 8 scenes with cameras facing forward, each containing about 20-62 images. The Mip-NeRF360 dataset contains different scenes where cameras are distributed evenly in 360 degrees in the 3D space, each containing about 100-300 images. The Tanks-and-Temples dataset is similar to the Mip-NeRF360 dataset in camera poses and scene scales but with more illumination and appearance changes.

\paragraph{Implementation Details.}
We initialize the scene regressor network using the pre-trained Spann3R model~\cite{DBLP:journals/corr/abs-2408-16061}. 
During training, we use the image resolution of $512 \times 512$ for all the datasets. We use a learning rate of $1e-5$ to finetune the scene regressor. 
We use $s_{\text{sim}}=0.3$ to reject edges when building the similarity graph.
We use DSAC~\cite{DBLP:conf/cvpr/BrachmannKNSMGR17,DBLP:journals/pami/BrachmannR22} to compute the camera poses for candidate image registration. Since DSAC supports only a single focal length for both the image x-axis and y-axis, we modify it to use different focal lengths for the image x-axis and y-axis. We set the threshold of inlier number $s_{\text{inlier}}$ to $5,000$ and the threshold of reprojection error to be within $6$ pixels in DSAC. For the seed initialization, we finetune the scene regressor by 500 iterations. During the incremental training, we finetune the scene regressor by 1,000 iterations on the LLFF dataset and 1,500 iterations on the Mip-NeRF360 dataset. For the novel view synthesis task, images are downsampled by 4 during training and inference.
\vspace{-2mm}

\begin{table*}[htbp]
    \centering

    \resizebox{1.0\textwidth}{!}{
    \begin{tabular}{l || c c c  c c c | c c c  c c c | c c c }
    \toprule
    
    \multirow{2}{*}{\textbf{Scenes}} &
    \multicolumn{3}{c}{\textbf{NeRF}~\cite{DBLP:conf/eccv/MildenhallSTBRN20}} &
    \multicolumn{3}{c}{\textbf{BARF}~\cite{DBLP:conf/iccv/LinM0L21}} $\ \ |$ & 
    \multicolumn{3}{c}{\textbf{3DGS}} &
    \multicolumn{3}{c}{\textbf{CF-3DGS}~\cite{DBLP:conf/cvpr/Fu0LKKE24}} $\ \ |$ & 
    \multicolumn{3}{c}{\textbf{$\text{Ours}$}} \\
    
    \cmidrule(r){2-4} \cmidrule(r){5-7} \cmidrule(r){8-10} \cmidrule(r){11-13} \cmidrule(r){14-16} 
   
    & \multirow{1}{*}{PSNR\ $\uparrow$}
    & \multirow{1}{*}{SSIM\ $\uparrow$} 
    & \multirow{1}{*}{LPIPS\ $\downarrow$} 
    
    & \multirow{1}{*}{PSNR\ $\uparrow$}
    & \multirow{1}{*}{SSIM\ $\uparrow$} 
    & \multirow{1}{*}{LPIPS\ $\downarrow$} 
    
    & \multirow{1}{*}{PSNR\ $\uparrow$}
    & \multirow{1}{*}{SSIM\ $\uparrow$} 
    & \multirow{1}{*}{LPIPS\ $\downarrow$} 

    & \multirow{1}{*}{PSNR\ $\uparrow$}
    & \multirow{1}{*}{SSIM\ $\uparrow$} 
    & \multirow{1}{*}{LPIPS\ $\downarrow$} 
   
    & \multirow{1}{*}{PSNR\ $\uparrow$}
    & \multirow{1}{*}{SSIM\ $\uparrow$} 
    & \multirow{1}{*}{LPIPS\ $\downarrow$} \\
   
    \midrule
   
    Fern
    & \cellcolor{yellow}23.72 & \cellcolor{yellow}0.733 & \cellcolor{yellow}0.262
    & \cellcolor{red}23.79 & 0.710 & 0.311 
    & 23.63 & \cellcolor{orange}0.794 & \cellcolor{orange}0.136
    & 17.35 & 0.494 & 0.428
    & \cellcolor{orange}23.77 & \cellcolor{red}0.797 & \cellcolor{red}0.114 \\
   
    Flower
    & 23.24 & 0.668 & 0.244
    & \cellcolor{yellow}23.37 & \cellcolor{yellow}0.698 & \cellcolor{yellow}0.211 
    & \cellcolor{red}26.91 & \cellcolor{red}0.829 & \cellcolor{red}0.096
    & 20.17 & 0.622 & 0.362
    & \cellcolor{orange}25.81 & \cellcolor{orange}0.812 & \cellcolor{orange}0.108 \\
   
    Fortress
    & 25.97 & 0.786 & 0.185
    & \cellcolor{yellow}29.08 & \cellcolor{yellow}0.823 & \cellcolor{yellow}0.132  
    & \cellcolor{red}29.93 & \cellcolor{red}0.880 & \cellcolor{red}0.078
    & 14.73 & 0.395 & 0.460
    & \cellcolor{orange}29.31 & \cellcolor{orange}0.868 & \cellcolor{orange}0.084 \\
   
    Horns
    & 20.35 & 0.624 & 0.421
    & \cellcolor{yellow}22.78 & \cellcolor{yellow}0.727 & \cellcolor{yellow}0.298 
    & \cellcolor{orange}26.02 & \cellcolor{orange}0.862 & \cellcolor{orange}0.121
    & 15.60 & 0.412 & 0.514
    & \cellcolor{red}26.67 & \cellcolor{red}0.882 & \cellcolor{red}0.093 \\

    Leaves
    & 15.33 & 0.306 & 0.526
    & \cellcolor{red}18.78 & \cellcolor{orange}0.537 & \cellcolor{yellow}0.353 
    & \cellcolor{orange}17.91 & \cellcolor{red}0.593 & \cellcolor{red}0.205
    & 15.38 & 0.416 & 0.398
    & \cellcolor{yellow}16.45 & \cellcolor{yellow}0.526 & \cellcolor{orange}0.270 \\

    Orchids
    & 17.34 & 0.518 & 0.307
    & \cellcolor{orange}19.45 & \cellcolor{yellow}0.574 & \cellcolor{yellow}0.291 
    & \cellcolor{yellow}18.98 & \cellcolor{orange}0.612 & \cellcolor{orange}0.159
    & 13.80 & 0.258 & 0.516 
    & \cellcolor{red}19.55 & \cellcolor{red}0.640 & \cellcolor{red}0.147 \\

    Room
    & \cellcolor{red}32.42 & \cellcolor{red}0.948 & \cellcolor{orange}0.080
    & \cellcolor{yellow}31.95 & \cellcolor{orange}0.940 & \cellcolor{yellow}0.099 
    & 28.96 & \cellcolor{yellow}0.927 & 0.115
    & 18.36 & 0.713 & 0.382 
    & \cellcolor{orange}32.37 & \cellcolor{red}0.948 & \cellcolor{red}0.073 \\

    Trex
    & 22.12 & 0.739 & 0.244
    & \cellcolor{yellow}22.55 & \cellcolor{yellow}0.767 & \cellcolor{yellow}0.206  
    & \cellcolor{orange}24.74 & \cellcolor{orange}0.881 & \cellcolor{orange}0.145
    & 16.76 & 0.522 & 0.434
    & \cellcolor{red}26.11 & \cellcolor{red}0.905 & \cellcolor{red}0.084 \\

    \bottomrule
    \end{tabular}
 }
    \vspace{-2mm}
    \caption{\textbf{Quantitative results of novel view synthesis on LLFF dataset}.
             $\uparrow$: higher is better, $\downarrow$: lower is better.
     }
    \label{table:quantitative_llff_nvs}
    \vspace{-3mm}
\end{table*}

\begin{figure*}[htbp]
    \centering
   
    \includegraphics[width=1.0\linewidth]{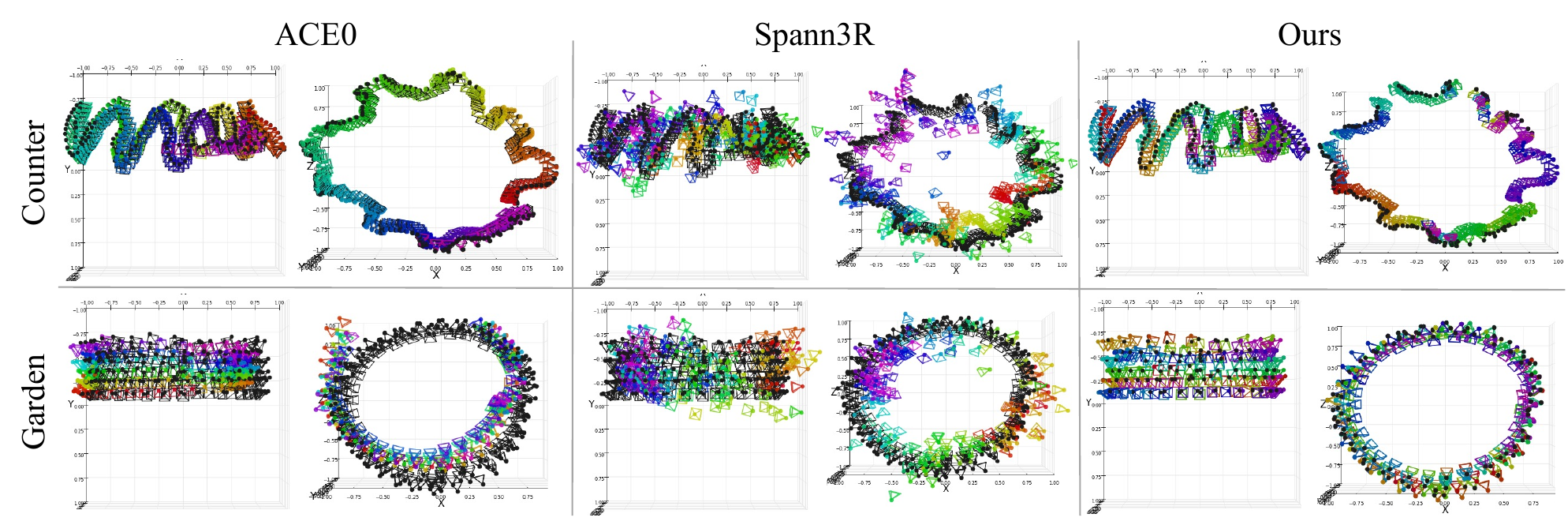}
    \vspace{-5mm}
    \caption{\textbf{Visualization of camera poses accuracy} on the MipNeRF360 dataset (Zoom in for best view).
     Black: pseudo-ground-truth camera poses obtained from COLMAP~\cite{DBLP:conf/cvpr/SchonbergerF16}. Colored: predicted camera poses.
     }
    \label{fig:mipnerf360_pose_cmp}
    \vspace{-5mm}
\end{figure*}

\begin{figure*}[htbp]
    \centering
    \subfloat {
        \includegraphics[width=0.95\linewidth]{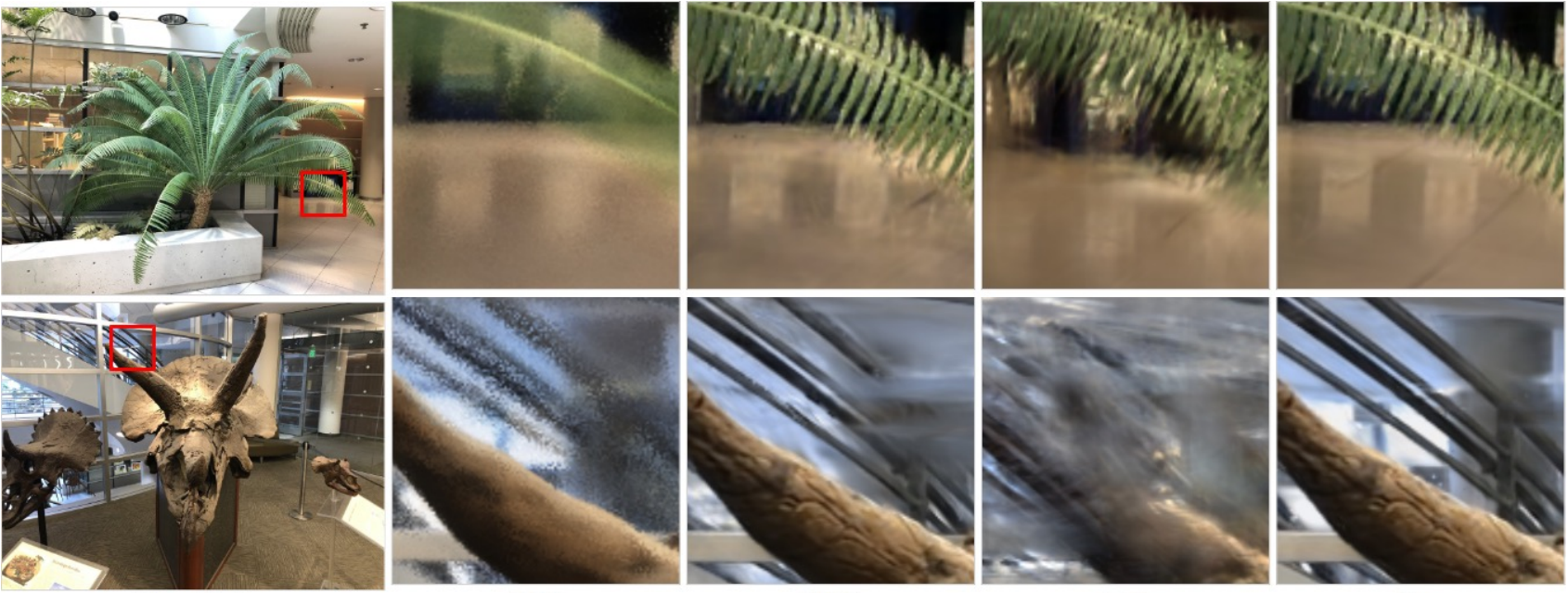}
     } \\
    \subfloat {
        \includegraphics[width=0.95\linewidth]{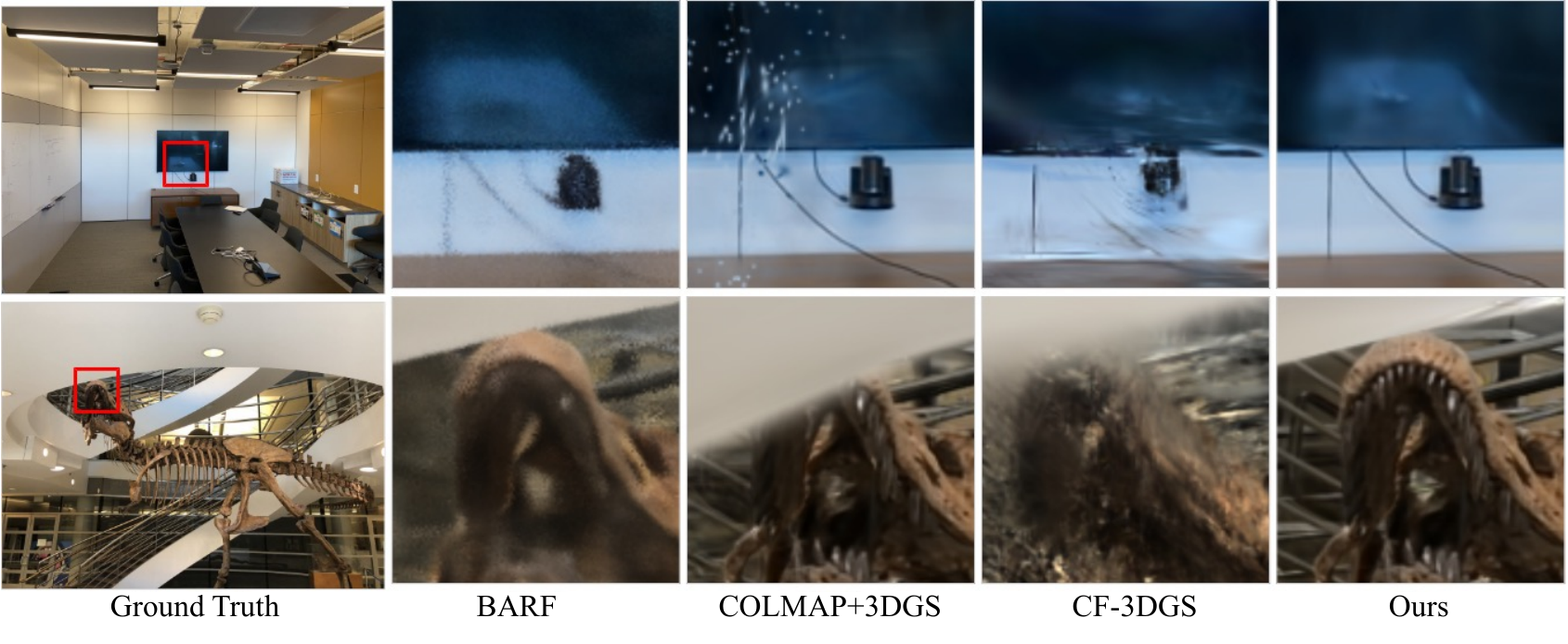}
     }
     \vspace{-3mm}
    \caption{\textbf{The qualitative results of novel view synthesis on LLFF forward-facing dataset~\cite{DBLP:journals/tog/MildenhallSCKRN19}}.}
    \label{fig:llff_render}
    \vspace{-5mm}
\end{figure*}

\paragraph{Results.}
We first present results on the \textbf{LLFF Dataset}. We compare our method with BARF~\cite{DBLP:conf/iccv/LinM0L21} and 
DBARF~\cite{DBLP:conf/cvpr/ChenL23}, which are NeRF-based pose-free methods. We also compare our method to the 
3DGS-based pose-free method CF-3DGS~\cite{DBLP:conf/cvpr/Fu0LKKE24}. Since InstantSplat~\cite{fan2024instantsplat} always reports an out-of-memory issue on the dataset, we do not compare to it. Besides pose-free NeRF/3DGS-based methods, we also compare with a scene regressor ACE0~\cite{DBLP:journals/corr/abs-2404-14351}. Note that Spann3R and ACE0 do not support novel view synthesis during this paper, and thus we only compare with it in terms of the camera pose accuracy. 
The quantitative results for camera pose accuracy are presented in Table~\ref{table:quantitative_llff_camera_pose_acc}. The unit for rotation error is degree, and the unit for translation error is dimensionless due to the loss of absolute scale in the ground-truth camera poses. 
The results show that the camera pose accuracy of our method consistently outperforms all other methods. BARF and DBARF are 
the second-best and third-best methods, suggesting that the pose-free methods in 3DGS are under-explored compared to the pose-free NeRF. While CF-3DGS claims to be COLMAP-free, it behaves badly on this dataset since the camera does not move on a smooth curve.
We also present the novel view synthesis results in Table~\ref{table:quantitative_llff_nvs}. Our method also renders the highest quality images than all other pose-free NeRF/3DGS methods. Surprisingly, our method even surpassed 3DGS trained with COLMAP pose on most scenes. This result suggests our camera pose can be even more accurate than COLMAP in those scenes.
We also present the qualitative comparison of the camera pose accuracy in Fig.~\ref{fig:llff_pose_cmp}. The ground-truth camera poses 
are rendered in black, and the predicted camera poses are rendered in rainbow colors. The qualitative results of novel view synthesis are given in Fig.~\ref{fig:llff_render}.

We also evaluate our method on the \textbf{Mip-NeRF360 dataset}. The Mip-NeRF360 dataset is more challenging than the LLFF dataset for pose-free NeRF/3DGS methods, where the latter contains only forward-facing views while the former is composed of cameras that rotate 360 degrees around the complex objects in the scene. Since BARF~\cite{DBLP:conf/iccv/LinM0L21} requires accurate initialization on non-forward-facing datasets and DBARF~\cite{DBLP:conf/cvpr/ChenL23} is designed mainly for scenes that are highly overlapped and have narrow baselines, we do not compare with these methods. We first present the camera pose accuracy in 
Table~\ref{table:quantitative_mipnerf360_camera_pose_acc}. We can observe that our method consistently outperforms ACE0 and Spann3R, while CF-3DGS always got an out-of-memory on the dataset, we mark its results by `-'. Some qualitative results of camera poses are shown in Fig.~\ref{fig:mipnerf360_pose_cmp}.

\begin{table}[htbp]
    \centering

    \resizebox{0.5\textwidth}{!}{
    \begin{tabular}{l || c c |  c c  c c | c c}
        \toprule
        
        \multirow{2}{*}{\textbf{Scenes}} &
        \multicolumn{2}{c}{\textbf{ACE0}~\cite{DBLP:journals/corr/abs-2404-14351}} $\ \ |$ &
        \multicolumn{2}{c}{\textbf{CF-3DGS}~\cite{DBLP:conf/cvpr/Fu0LKKE24}} &
        \multicolumn{2}{c}{\textbf{Spann3R}~\cite{DBLP:journals/corr/abs-2408-16061}} $\ \ |$ &
        \multicolumn{2}{c}{\textbf{$\text{Ours}$}} \\
        
        \cmidrule(r){2-3} \cmidrule(r){4-5} \cmidrule(r){6-7} \cmidrule(r){8-9}
       
        & \multirow{1}{*}{$\Delta \mathbf{R}$}
        & \multirow{1}{*}{$\Delta \mathbf{t}$} 
        
        & \multirow{1}{*}{$\Delta \mathbf{R}$}
        & \multirow{1}{*}{$\Delta \mathbf{t}$} 
       
        & \multirow{1}{*}{$\Delta \mathbf{R}$}
        & \multirow{1}{*}{$\Delta \mathbf{t}$} 
    
        & \multirow{1}{*}{$\Delta \mathbf{R}$}
        & \multirow{1}{*}{$\Delta \mathbf{t}$}  \\
       
        \midrule
       
        Bicycle 
        & \cellcolor{orange}4.56 & \cellcolor{orange}0.052  
        & - & -  
        & \cellcolor{yellow}10.79 & \cellcolor{yellow}0.212  
        & \cellcolor{red}0.035 & \cellcolor{red}0.005 
        \\
        
        Counter 
        & \cellcolor{orange}1.76 & \cellcolor{orange}0.017  
        & - & -  
        & \cellcolor{yellow}11.16 & \cellcolor{yellow}0.226  
        & \cellcolor{red}0.029 & \cellcolor{red}0.002 
        \\
       
        Garden  
        & \cellcolor{yellow}22.66 & \cellcolor{yellow}0.286  
        & - & -  
        & \cellcolor{orange}15.65 & \cellcolor{orange}0.279  
        & \cellcolor{red}0.028 & \cellcolor{red}0.002  
        \\
    
    
        Kitchen  
        & \cellcolor{orange}2.65 & \cellcolor{orange}0.044 
        & - & -  
        & \cellcolor{yellow}9.60 & \cellcolor{yellow}0.171  
        & \cellcolor{red}0.052 & \cellcolor{red}0.008   
        \\

        \bottomrule
    \end{tabular}
    }
    \vspace{-2mm}
    \caption{\textbf{Quantitative results of camera pose accuracy on MipNeRF360 dataset}.
     The \colorbox{red}{red}, \colorbox{orange}{orange} and \colorbox{yellow}{yellow} colors respectively 
     denote the best, the second best, and the third best results.
     }
    \label{table:quantitative_mipnerf360_camera_pose_acc}
    \vspace{-2mm}
\end{table}

\begin{table}[htbp]
    \centering

    \resizebox{0.5\textwidth}{!}{
    \begin{tabular}{l  c c c  c c c}
        \toprule
        
        \multirow{2}{*}{\textbf{Scenes}} &
        \multicolumn{3}{c}{\textbf{COLMAP+3DGS}} &
        \multicolumn{2}{c}{\textbf{$\text{Ours}$}} \\
        
        \cmidrule(r){2-4} \cmidrule(r){5-7} 
       
        & \multirow{1}{*}{PSNR\ $\uparrow$}
        & \multirow{1}{*}{SSIM\ $\uparrow$} 
        & \multirow{1}{*}{LPIPS\ $\downarrow$} 
        
        & \multirow{1}{*}{PSNR\ $\uparrow$}
        & \multirow{1}{*}{SSIM\ $\uparrow$} 
        & \multirow{1}{*}{LPIPS\ $\downarrow$}   \\
       
        \midrule
       
        Bicycle 
        & 22.92 & 0.695 & 0.201
        & \cellcolor{red}23.10 & \cellcolor{red}0.707 & \cellcolor{red}0.201
        \\
        
        Counter
        & \cellcolor{red}27.64 & \cellcolor{red}0.878 & \cellcolor{red}0.122
        & 26.87 & 0.873 & 0.124
        \\
       
        Garden
        & 24.83 & 0.769 & 0.153
        & \cellcolor{red}25.47 & \cellcolor{red}0.839 & \cellcolor{red}0.107
        \\
        
        Kitchen
        & \cellcolor{red}30.93 & \cellcolor{red}0.932 & \cellcolor{red}0.054
        & 29.67 & 0.925 & 0.061
        \\
       
        \bottomrule
    \end{tabular}
    }
    \vspace{-2mm}
    \caption{\textbf{Quantitative results of novel view synthesis on MipNeRF360 dataset}.
        $\uparrow$: higher is better, $\downarrow$: lower is better.
    }
    \label{table:quantitative_mipnerf360_nvs}
    \vspace{-6mm}
\end{table}

   
   


We further evaluate our method on the \textbf{Tanks-and-Temples dataset}. We present the final novel view synthesis results in Table~\ref{table:quantitative_tnt_nvs}. The higher image rendering quality of our method in Table~\ref{table:quantitative_tnt_nvs} suggests that our camera poses can be more accurate than COLMAP. See our supplementary for more qualitative and quantitative results on this dataset.

\begin{table}[htbp]
    \centering

    \resizebox{0.5\textwidth}{!}{
    \begin{tabular}{l  c c c  c c c}
        \toprule
        
        \multirow{2}{*}{\textbf{Scenes}} &
        \multicolumn{3}{c}{\textbf{COLMAP+3DGS}} &
        \multicolumn{2}{c}{\textbf{$\text{Ours}$}} \\
        
        \cmidrule(r){2-4} \cmidrule(r){5-7} 
       
        & \multirow{1}{*}{PSNR\ $\uparrow$}
        & \multirow{1}{*}{SSIM\ $\uparrow$} 
        & \multirow{1}{*}{LPIPS\ $\downarrow$} 
        
        & \multirow{1}{*}{PSNR\ $\uparrow$}
        & \multirow{1}{*}{SSIM\ $\uparrow$} 
        & \multirow{1}{*}{LPIPS\ $\downarrow$}   \\
       
        \midrule
       
        Family 
        & 21.92 & 0.733 & 0.177
        & \cellcolor{red}23.03 & \cellcolor{red}0.770 & \cellcolor{red}0.144
        \\
        
        Francis
        & 26.09 & 0.830 & 0.214
        & \cellcolor{red}26.79 & \cellcolor{red}0.842 & \cellcolor{red}0.186
        \\
       
        Ignatius
        & 20.90 & 0.606 & 0.281
        & \cellcolor{red}21.95 & \cellcolor{red}0.665 & \cellcolor{red}0.234
        \\

        Train
        & 19.68 & 0.646 & 0.277
        & \cellcolor{red}20.59 & \cellcolor{red}0.673 & \cellcolor{red}0.252
        \\
       
        \bottomrule
    \end{tabular}
    }
    \vspace{-2mm}
    \caption{\textbf{Quantitative results of novel view synthesis on Tanks-and-Temples dataset}.
        $\uparrow$: higher is better, $\downarrow$: lower is better.
    }
    \label{table:quantitative_tnt_nvs}
    \vspace{-5mm}
\end{table}

\paragraph{Ablation Study.} We ablate the effectiveness of our incremental training step and camera pose refinement step in Table~\ref{table:ablation_study_llff_camera_pose_acc}. We can see that with the refinement step, the camera pose accuracy is improved significantly.
Although our proposed method is initialized from the pre-trained Spann3R model, the convergence of the refinement step can still fail in cases where the derived camera poses are grossly erroneous.
We observe that camera pose accuracy is improved with our incremental training pipeline ($\text{Ours}_{\text{coarse}}$ \emph{v.s.} Spann3R), which provides much better initial values than Spann3R. See our supplementary for more ablation results.

   

\begin{table}[htbp]
    \centering

    \resizebox{0.48\textwidth}{!}{

    \begin{tabular}{l  c c  c c  c c}
        \toprule
        
        \multirow{2}{*}{\textbf{Scenes}} &
        \multicolumn{2}{c}{\textbf{Spann3R}~\cite{DBLP:journals/corr/abs-2408-16061}} &
        \multicolumn{2}{c}{\textbf{$\text{Ours}_{\text{coarse}}$}} &
        \multicolumn{2}{c}{\textbf{$\text{Ours}_{\text{refine}}$}} \\
        
        \cmidrule(r){2-3} \cmidrule(r){4-5} \cmidrule(r){6-7} 
       
        & \multirow{1}{*}{$\Delta \mathbf{R}$}
        & \multirow{1}{*}{$\Delta \mathbf{t}$} 
        
        & \multirow{1}{*}{$\Delta \mathbf{R}$}
        & \multirow{1}{*}{$\Delta \mathbf{t}$} 
    
        & \multirow{1}{*}{$\Delta \mathbf{R}$}
        & \multirow{1}{*}{$\Delta \mathbf{t}$}  \\
       
        \midrule
       
        Fern
        &  \cellcolor{yellow}39.03 &  \cellcolor{yellow}0.767
        &  \cellcolor{orange}01.30 &  \cellcolor{orange}0.125 
        &  \cellcolor{red}0.26 & \cellcolor{red}0.005 \\

    
        Flower
        & \cellcolor{orange}11.91 & \cellcolor{orange}0.285 
        & \cellcolor{yellow}16.53 & \cellcolor{yellow}0.609
        & \cellcolor{red}0.52 & \cellcolor{red}0.011 \\
       
        
        Fortress
        & \cellcolor{yellow}08.31 & \cellcolor{yellow}0.152 
        & \cellcolor{orange}06.07 & \cellcolor{orange}0.127
        & \cellcolor{red}0.04 & \cellcolor{red}0.002 \\
    
       
        Horns
        & \cellcolor{orange}06.98 & \cellcolor{yellow}0.349 
        & \cellcolor{yellow}14.23 & \cellcolor{orange}0.145
        & \cellcolor{red}0.03 & \cellcolor{red}0.001 \\
    
    
        Leaves 
        & \cellcolor{yellow}44.09 & \cellcolor{yellow}0.801 
        & \cellcolor{orange}18.24 &\cellcolor{orange} 0.187
        & \cellcolor{red}0.22 & \cellcolor{red}0.006 \\
    
    
        Orchids
        & \cellcolor{yellow}09.77 & \cellcolor{yellow}0.256
        & \cellcolor{orange}07.22 & \cellcolor{orange}0.255
        & \cellcolor{red}0.24 & \cellcolor{red}0.006 \\
    
    
        Room
        & \cellcolor{orange}07.48 & \cellcolor{yellow}0.513
        & \cellcolor{yellow}10.22 & \cellcolor{orange}0.180 
        &  \cellcolor{red}0.03 & \cellcolor{red}0.001 \\
    
    
        Trex
        & \cellcolor{yellow}32.39 & \cellcolor{yellow}0.758
        & \cellcolor{orange}07.76 & \cellcolor{orange}0.210 
        & \cellcolor{red}0.03 & \cellcolor{red}0.010 \\
       
        \bottomrule
    \end{tabular}
    }
    \vspace{-3mm}
    \caption{\textbf{Ablation study of camera pose accuracy}.
    }
    \label{table:ablation_study_llff_camera_pose_acc}
    \vspace{-6mm}
\end{table}

\section{Conclusion}
\label{sec:conclusion}

In this paper, we propose ZeroGS to reconstruct neural scenes from unposed images. Our method adopts a pretrained 3D foundation model as a scene regressor and leverages its learned geometry priors to ease the task of pose-free 3DGS training. Based on the learned geometry, we obtain coarse camera poses by RANSAC and PnP solver and refine it with a point-to-camera ray consistency loss. Our training pipeline incrementally registers the image batch into a training buffer and progressively finetunes the model in a self-supervised manner. Our method surpassed state-of-the-art pose-free NeRF/3DGS methods on the LLFF, MipNeRF360, and Tanks-and-Temples datasets and comparable or even outperforms 3DGS trained with COLMAP poses.

{
    \clearpage
    \newpage
    \small
    \bibliographystyle{ieeenat_fullname}
    \bibliography{main}

\begin{thebibliography}{67}
\providecommand{\natexlab}[1]{#1}
\providecommand{\url}[1]{\texttt{#1}}
\expandafter\ifx\csname urlstyle\endcsname\relax
  \providecommand{\doi}[1]{doi: #1}\else
  \providecommand{\doi}{doi: \begingroup \urlstyle{rm}\Url}\fi

\bibitem[Arandjelovic et~al.(2016)Arandjelovic, Gron{\'{a}}t, Torii, Pajdla,
  and Sivic]{DBLP:conf/cvpr/ArandjelovicGTP16}
Relja Arandjelovic, Petr Gron{\'{a}}t, Akihiko Torii, Tom{\'{a}}s Pajdla, and
  Josef Sivic.
\newblock Netvlad: {CNN} architecture for weakly supervised place recognition.
\newblock In \emph{{IEEE} Conference on Computer Vision and Pattern
  Recognition}, pages 5297--5307, 2016.

\bibitem[Barron et~al.(2021)Barron, Mildenhall, Tancik, Hedman,
  Martin{-}Brualla, and Srinivasan]{DBLP:conf/iccv/BarronMTHMS21}
Jonathan~T. Barron, Ben Mildenhall, Matthew Tancik, Peter Hedman, Ricardo
  Martin{-}Brualla, and Pratul~P. Srinivasan.
\newblock Mip-nerf: {A} multiscale representation for anti-aliasing neural
  radiance fields.
\newblock In \emph{{IEEE/CVF} International Conference on Computer Vision},
  pages 5835--5844, 2021.

\bibitem[Barron et~al.(2022)Barron, Mildenhall, Verbin, Srinivasan, and
  Hedman]{DBLP:conf/cvpr/BarronMVSH22}
Jonathan~T. Barron, Ben Mildenhall, Dor Verbin, Pratul~P. Srinivasan, and Peter
  Hedman.
\newblock Mip-nerf 360: Unbounded anti-aliased neural radiance fields.
\newblock In \emph{{IEEE/CVF} Conference on Computer Vision and Pattern
  Recognition}, pages 5460--5469, 2022.

\bibitem[Barron et~al.(2023)Barron, Mildenhall, Verbin, Srinivasan, and
  Hedman]{DBLP:conf/iccv/BarronMVSH23}
Jonathan~T. Barron, Ben Mildenhall, Dor Verbin, Pratul~P. Srinivasan, and Peter
  Hedman.
\newblock Zip-nerf: Anti-aliased grid-based neural radiance fields.
\newblock In \emph{{IEEE/CVF} International Conference on Computer Vision},
  pages 19640--19648, 2023.

\bibitem[Bian et~al.(2023)Bian, Wang, Li, and Bian]{DBLP:conf/cvpr/BianWLB23}
Wenjing Bian, Zirui Wang, Kejie Li, and Jia{-}Wang Bian.
\newblock Nope-nerf: Optimising neural radiance field with no pose prior.
\newblock In \emph{{IEEE/CVF} Conference on Computer Vision and Pattern
  Recognition}, pages 4160--4169, 2023.

\bibitem[Brachmann and Rother(2022)]{DBLP:journals/pami/BrachmannR22}
Eric Brachmann and Carsten Rother.
\newblock Visual camera re-localization from {RGB} and {RGB-D} images using
  {DSAC}.
\newblock \emph{{IEEE} Trans. Pattern Anal. Mach. Intell.}, 44\penalty0
  (9):\penalty0 5847--5865, 2022.

\bibitem[Brachmann et~al.(2017)Brachmann, Krull, Nowozin, Shotton, Michel,
  Gumhold, and Rother]{DBLP:conf/cvpr/BrachmannKNSMGR17}
Eric Brachmann, Alexander Krull, Sebastian Nowozin, Jamie Shotton, Frank
  Michel, Stefan Gumhold, and Carsten Rother.
\newblock {DSAC} - differentiable {RANSAC} for camera localization.
\newblock In \emph{{IEEE} Conference on Computer Vision and Pattern
  Recognition}, pages 2492--2500, 2017.

\bibitem[Brachmann et~al.(2023)Brachmann, Cavallari, and
  Prisacariu]{DBLP:conf/cvpr/BrachmannCP23}
Eric Brachmann, Tommaso Cavallari, and Victor~Adrian Prisacariu.
\newblock Accelerated coordinate encoding: Learning to relocalize in minutes
  using {RGB} and poses.
\newblock In \emph{{IEEE/CVF} Conference on Computer Vision and Pattern
  Recognition}, pages 5044--5053, 2023.

\bibitem[Brachmann et~al.(2024)Brachmann, Wynn, Chen, Cavallari, Monszpart,
  Turmukhambetov, and Prisacariu]{DBLP:journals/corr/abs-2404-14351}
Eric Brachmann, Jamie Wynn, Shuai Chen, Tommaso Cavallari, {\'{A}}ron
  Monszpart, Daniyar Turmukhambetov, and Victor~Adrian Prisacariu.
\newblock Scene coordinate reconstruction: Posing of image collections via
  incremental learning of a relocalizer.
\newblock \emph{CoRR}, abs/2404.14351, 2024.

\bibitem[Chen et~al.(2021)Chen, Xu, Zhao, Zhang, Xiang, Yu, and
  Su]{DBLP:conf/iccv/ChenXZZXY021}
Anpei Chen, Zexiang Xu, Fuqiang Zhao, Xiaoshuai Zhang, Fanbo Xiang, Jingyi Yu,
  and Hao Su.
\newblock Mvsnerf: Fast generalizable radiance field reconstruction from
  multi-view stereo.
\newblock In \emph{{IEEE/CVF} International Conference on Computer Vision},
  pages 14104--14113, 2021.

\bibitem[Chen et~al.(2022)Chen, Xu, Geiger, Yu, and
  Su]{DBLP:conf/eccv/ChenXGYS22}
Anpei Chen, Zexiang Xu, Andreas Geiger, Jingyi Yu, and Hao Su.
\newblock Tensorf: Tensorial radiance fields.
\newblock In \emph{Computer Vision - {ECCV} 2022 - 17th European Conference},
  pages 333--350, 2022.

\bibitem[Chen and Lee(2023{\natexlab{a}})]{DBLP:conf/cvpr/ChenL23}
Yu Chen and Gim~Hee Lee.
\newblock {DBARF:} deep bundle-adjusting generalizable neural radiance fields.
\newblock In \emph{{IEEE/CVF} Conference on Computer Vision and Pattern
  Recognition}, pages 24--34, 2023{\natexlab{a}}.

\bibitem[Chen and Lee(2023{\natexlab{b}})]{DBLP:conf/iccv/ChenL23}
Yu Chen and Gim~Hee Lee.
\newblock Dreg-nerf: Deep registration for neural radiance fields.
\newblock In \emph{{IEEE/CVF} International Conference on Computer Vision},
  pages 22646--22656, 2023{\natexlab{b}}.

\bibitem[Chen and Lee(2024)]{DBLP:journals/corr/abs-2405-13943}
Yu Chen and Gim~Hee Lee.
\newblock Dogaussian: Distributed-oriented gaussian splatting for large-scale
  3d reconstruction via gaussian consensus.
\newblock \emph{CoRR}, abs/2405.13943, 2024.

\bibitem[Chen et~al.(2023)Chen, Chen, Wang, Zhang, Guo, Shan, and
  Wang]{DBLP:conf/cvpr/ChenCWZ0S023}
Yue Chen, Xingyu Chen, Xuan Wang, Qi Zhang, Yu Guo, Ying Shan, and Fei Wang.
\newblock Local-to-global registration for bundle-adjusting neural radiance
  fields.
\newblock In \emph{{IEEE/CVF} Conference on Computer Vision and Pattern
  Recognition}, pages 8264--8273, 2023.

\bibitem[Chng et~al.(2022)Chng, Ramasinghe, Sherrah, and
  Lucey]{DBLP:journals/corr/abs-2204-05735}
Shin{-}Fang Chng, Sameera Ramasinghe, Jamie Sherrah, and Simon Lucey.
\newblock {GARF:} gaussian activated radiance fields for high fidelity
  reconstruction and pose estimation.
\newblock \emph{CoRR}, abs/2204.05735, 2022.

\bibitem[Dosovitskiy et~al.(2021)Dosovitskiy, Beyer, Kolesnikov, Weissenborn,
  Zhai, Unterthiner, Dehghani, Minderer, Heigold, Gelly, Uszkoreit, and
  Houlsby]{DBLP:conf/iclr/DosovitskiyB0WZ21}
Alexey Dosovitskiy, Lucas Beyer, Alexander Kolesnikov, Dirk Weissenborn,
  Xiaohua Zhai, Thomas Unterthiner, Mostafa Dehghani, Matthias Minderer, Georg
  Heigold, Sylvain Gelly, Jakob Uszkoreit, and Neil Houlsby.
\newblock An image is worth 16x16 words: Transformers for image recognition at
  scale.
\newblock In \emph{9th International Conference on Learning Representations,
  {ICLR} 2021}, 2021.

\bibitem[Fan et~al.(2024)Fan, Cong, Wen, Wang, Zhang, Ding, Xu, Ivanovic,
  Pavone, Pavlakos, Wang, and Wang]{fan2024instantsplat}
Zhiwen Fan, Wenyan Cong, Kairun Wen, Kevin Wang, Jian Zhang, Xinghao Ding,
  Danfei Xu, Boris Ivanovic, Marco Pavone, Georgios Pavlakos, Zhangyang Wang,
  and Yue Wang.
\newblock Instantsplat: Unbounded sparse-view pose-free gaussian splatting in
  40 seconds, 2024.

\bibitem[Fridovich{-}Keil et~al.(2022)Fridovich{-}Keil, Yu, Tancik, Chen,
  Recht, and Kanazawa]{DBLP:conf/cvpr/Fridovich-KeilY22}
Sara Fridovich{-}Keil, Alex Yu, Matthew Tancik, Qinhong Chen, Benjamin Recht,
  and Angjoo Kanazawa.
\newblock Plenoxels: Radiance fields without neural networks.
\newblock In \emph{{IEEE/CVF} Conference on Computer Vision and Pattern
  Recognition}, pages 5491--5500, 2022.

\bibitem[Fu et~al.(2024{\natexlab{a}})Fu, Yu, Li, and
  Zhang]{DBLP:journals/tmm/FuYLZ24}
Hongyu Fu, Xin Yu, Lincheng Li, and Li Zhang.
\newblock {CBARF:} cascaded bundle-adjusting neural radiance fields from
  imperfect camera poses.
\newblock \emph{{IEEE} Trans. Multim.}, 26:\penalty0 9304--9315,
  2024{\natexlab{a}}.

\bibitem[Fu et~al.(2024{\natexlab{b}})Fu, Wang, Liu, Kulkarni, Kautz, and
  Efros]{DBLP:conf/cvpr/Fu0LKKE24}
Yang Fu, Xiaolong Wang, Sifei Liu, Amey Kulkarni, Jan Kautz, and Alexei~A.
  Efros.
\newblock Colmap-free 3d gaussian splatting.
\newblock In \emph{{IEEE/CVF} Conference on Computer Vision and Pattern
  Recognition}, pages 20796--20805, 2024{\natexlab{b}}.

\bibitem[Goli et~al.(2023)Goli, Rebain, Sabour, Garg, and
  Tagliasacchi]{DBLP:conf/icra/GoliRSGT23}
Lily Goli, Daniel Rebain, Sara Sabour, Animesh Garg, and Andrea Tagliasacchi.
\newblock nerf2nerf: Pairwise registration of neural radiance fields.
\newblock In \emph{{IEEE} International Conference on Robotics and Automation},
  pages 9354--9361, 2023.

\bibitem[Govindu(2004)]{DBLP:conf/cvpr/Govindu04}
Venu~Madhav Govindu.
\newblock Lie-algebraic averaging for globally consistent motion estimation.
\newblock In \emph{2004 {IEEE} Computer Society Conference on Computer Vision
  and Pattern Recognition}, pages 684--691, 2004.

\bibitem[Gu{\'{e}}don and Lepetit(2024)]{DBLP:conf/cvpr/GuedonL24}
Antoine Gu{\'{e}}don and Vincent Lepetit.
\newblock Sugar: Surface-aligned gaussian splatting for efficient 3d mesh
  reconstruction and high-quality mesh rendering.
\newblock In \emph{{IEEE/CVF} Conference on Computer Vision and Pattern
  Recognition}, pages 5354--5363, 2024.

\bibitem[Huang et~al.(2024)Huang, Yu, Chen, Geiger, and
  Gao]{DBLP:conf/siggraph/HuangYC0G24}
Binbin Huang, Zehao Yu, Anpei Chen, Andreas Geiger, and Shenghua Gao.
\newblock 2d gaussian splatting for geometrically accurate radiance fields.
\newblock In \emph{{ACM} {SIGGRAPH} 2024 Conference Papers}, page~32, 2024.

\bibitem[Jain et~al.(2022)Jain, Kumar, and
  Gool]{DBLP:journals/corr/abs-2210-04233}
Nishant Jain, Suryansh Kumar, and Luc~Van Gool.
\newblock Robustifying the multi-scale representation of neural radiance
  fields.
\newblock \emph{CoRR}, abs/2210.04233, 2022.

\bibitem[Johari et~al.(2022)Johari, Lepoittevin, and
  Fleuret]{DBLP:conf/cvpr/JohariLF22}
Mohammad~Mahdi Johari, Yann Lepoittevin, and Fran{\c{c}}ois Fleuret.
\newblock Geonerf: Generalizing nerf with geometry priors.
\newblock In \emph{{IEEE/CVF} Conference on Computer Vision and Pattern
  Recognition}, pages 18344--18347, 2022.

\bibitem[Kerbl et~al.(2023)Kerbl, Kopanas, Leimk{\"{u}}hler, and
  Drettakis]{DBLP:journals/tog/KerblKLD23}
Bernhard Kerbl, Georgios Kopanas, Thomas Leimk{\"{u}}hler, and George
  Drettakis.
\newblock 3d gaussian splatting for real-time radiance field rendering.
\newblock \emph{{ACM} Trans. Graph.}, 42\penalty0 (4):\penalty0 139:1--139:14,
  2023.

\bibitem[Kerbl et~al.(2024)Kerbl, Meuleman, Kopanas, Wimmer, Lanvin, and
  Drettakis]{DBLP:journals/tog/KerblMKWLD24}
Bernhard Kerbl, Andreas Meuleman, Georgios Kopanas, Michael Wimmer, Alexandre
  Lanvin, and George Drettakis.
\newblock A hierarchical 3d gaussian representation for real-time rendering of
  very large datasets.
\newblock \emph{{ACM} Trans. Graph.}, 43\penalty0 (4):\penalty0 62:1--62:15,
  2024.

\bibitem[Kingma and Ba(2015)]{DBLP:journals/corr/KingmaB14}
Diederik~P. Kingma and Jimmy Ba.
\newblock Adam: {A} method for stochastic optimization.
\newblock In \emph{3rd International Conference on Learning Representations},
  2015.

\bibitem[Knapitsch et~al.(2017)Knapitsch, Park, Zhou, and
  Koltun]{Knapitsch2017}
Arno Knapitsch, Jaesik Park, Qian-Yi Zhou, and Vladlen Koltun.
\newblock Tanks and temples: Benchmarking large-scale scene reconstruction.
\newblock \emph{ACM Transactions on Graphics}, 36\penalty0 (4), 2017.

\bibitem[Liang et~al.(2024)Liang, Zhang, Hu, Feng, Zhu, and
  Jia]{DBLP:journals/corr/abs-2403-11056}
Zhihao Liang, Qi Zhang, Wenbo Hu, Ying Feng, Lei Zhu, and Kui Jia.
\newblock Analytic-splatting: Anti-aliased 3d gaussian splatting via analytic
  integration.
\newblock \emph{CoRR}, abs/2403.11056, 2024.

\bibitem[Lin et~al.(2021)Lin, Ma, Torralba, and Lucey]{DBLP:conf/iccv/LinM0L21}
Chen{-}Hsuan Lin, Wei{-}Chiu Ma, Antonio Torralba, and Simon Lucey.
\newblock {BARF:} bundle-adjusting neural radiance fields.
\newblock In \emph{{IEEE/CVF} International Conference on Computer Vision},
  pages 5721--5731, 2021.

\bibitem[Lin et~al.(2024)Lin, Li, Tang, Liu, Liu, Liu, Lu, Wu, Xu, Yan, and
  Yang]{DBLP:conf/cvpr/LinLTLLLLWXYY24}
Jiaqi Lin, Zhihao Li, Xiao Tang, Jianzhuang Liu, Shiyong Liu, Jiayue Liu,
  Yangdi Lu, Xiaofei Wu, Songcen Xu, Youliang Yan, and Wenming Yang.
\newblock Vastgaussian: Vast 3d gaussians for large scene reconstruction.
\newblock In \emph{{IEEE/CVF} Conference on Computer Vision and Pattern
  Recognition}, pages 5166--5175, 2024.

\bibitem[Liu et~al.(2020)Liu, Gu, Lin, Chua, and
  Theobalt]{DBLP:conf/nips/LiuGLCT20}
Lingjie Liu, Jiatao Gu, Kyaw~Zaw Lin, Tat{-}Seng Chua, and Christian Theobalt.
\newblock Neural sparse voxel fields.
\newblock In \emph{Advances in Neural Information Processing Systems 33}, 2020.

\bibitem[Liu et~al.(2022)Liu, Peng, Liu, Wang, Wang, Theobalt, Zhou, and
  Wang]{DBLP:conf/cvpr/LiuPLWWTZW22}
Yuan Liu, Sida Peng, Lingjie Liu, Qianqian Wang, Peng Wang, Christian Theobalt,
  Xiaowei Zhou, and Wenping Wang.
\newblock Neural rays for occlusion-aware image-based rendering.
\newblock In \emph{{IEEE/CVF} Conference on Computer Vision and Pattern
  Recognition}, pages 7814--7823, 2022.

\bibitem[Liu et~al.(2024)Liu, Guan, Luo, Fan, Peng, and
  Zhang]{DBLP:journals/corr/abs-2404-01133}
Yang Liu, He Guan, Chuanchen Luo, Lue Fan, Junran Peng, and Zhaoxiang Zhang.
\newblock Citygaussian: Real-time high-quality large-scale scene rendering with
  gaussians.
\newblock \emph{CoRR}, abs/2404.01133, 2024.

\bibitem[Lu et~al.(2024)Lu, Yu, Xu, Xiangli, Wang, Lin, and
  Dai]{DBLP:conf/cvpr/0005YXX0L024}
Tao Lu, Mulin Yu, Linning Xu, Yuanbo Xiangli, Limin Wang, Dahua Lin, and Bo
  Dai.
\newblock Scaffold-gs: Structured 3d gaussians for view-adaptive rendering.
\newblock In \emph{{IEEE/CVF} Conference on Computer Vision and Pattern
  Recognition}, pages 20654--20664, 2024.

\bibitem[Mi and Xu(2023)]{DBLP:conf/iclr/Mi023}
Zhenxing Mi and Dan Xu.
\newblock Switch-nerf: Learning scene decomposition with mixture of experts for
  large-scale neural radiance fields.
\newblock In \emph{The Eleventh International Conference on Learning
  Representations}, 2023.

\bibitem[Mildenhall et~al.()Mildenhall, Srinivasan, Tancik, Barron,
  Ramamoorthi, and Ng]{DBLP:conf/eccv/MildenhallSTBRN20}
Ben Mildenhall, Pratul~P. Srinivasan, Matthew Tancik, Jonathan~T. Barron, Ravi
  Ramamoorthi, and Ren Ng.
\newblock Nerf: Representing scenes as neural radiance fields for view
  synthesis.
\newblock In \emph{Computer Vision - {ECCV} 2020 - 16th European Conference,
  Glasgow}, pages 405--421.

\bibitem[Mildenhall et~al.(2019)Mildenhall, Srinivasan, Cayon, Kalantari,
  Ramamoorthi, Ng, and Kar]{DBLP:journals/tog/MildenhallSCKRN19}
Ben Mildenhall, Pratul~P. Srinivasan, Rodrigo~Ortiz Cayon, Nima~Khademi
  Kalantari, Ravi Ramamoorthi, Ren Ng, and Abhishek Kar.
\newblock Local light field fusion: practical view synthesis with prescriptive
  sampling guidelines.
\newblock \emph{{ACM} Trans. Graph.}, 38\penalty0 (4):\penalty0 29:1--29:14,
  2019.

\bibitem[M{\"{u}}ller et~al.(2022)M{\"{u}}ller, Evans, Schied, and
  Keller]{DBLP:journals/tog/MullerESK22}
Thomas M{\"{u}}ller, Alex Evans, Christoph Schied, and Alexander Keller.
\newblock Instant neural graphics primitives with a multiresolution hash
  encoding.
\newblock \emph{{ACM} Trans. Graph.}, 41\penalty0 (4):\penalty0 102:1--102:15,
  2022.

\bibitem[Purkait et~al.(2020)Purkait, Chin, and
  Reid]{DBLP:conf/eccv/PurkaitCR20}
Pulak Purkait, Tat{-}Jun Chin, and Ian Reid.
\newblock Neurora: Neural robust rotation averaging.
\newblock In \emph{Computer Vision - {ECCV} 2020 - 16th European Conference},
  pages 137--154, 2020.

\bibitem[Ranftl et~al.(2021)Ranftl, Bochkovskiy, and
  Koltun]{DBLP:conf/iccv/RanftlBK21}
Ren{\'{e}} Ranftl, Alexey Bochkovskiy, and Vladlen Koltun.
\newblock Vision transformers for dense prediction.
\newblock In \emph{{IEEE/CVF} International Conference on Computer Vision},
  pages 12159--12168, 2021.

\bibitem[Rematas et~al.(2022)Rematas, Liu, Srinivasan, Barron, Tagliasacchi,
  Funkhouser, and Ferrari]{DBLP:conf/cvpr/RematasLSBTFF22}
Konstantinos Rematas, Andrew Liu, Pratul~P. Srinivasan, Jonathan~T. Barron,
  Andrea Tagliasacchi, Thomas~A. Funkhouser, and Vittorio Ferrari.
\newblock Urban radiance fields.
\newblock In \emph{{IEEE/CVF} Conference on Computer Vision and Pattern
  Recognition}, pages 12922--12932, 2022.

\bibitem[Ren et~al.(2024)Ren, Jiang, Lu, Yu, Xu, Ni, and Dai]{octreegs}
Kerui Ren, Lihan Jiang, Tao Lu, Mulin Yu, Linning Xu, Zhangkai Ni, and Bo Dai.
\newblock Octree-gs: Towards consistent real-time rendering with lod-structured
  3d gaussians, 2024.

\bibitem[Sch{\"{o}}nberger and Frahm(2016)]{DBLP:conf/cvpr/SchonbergerF16}
Johannes~L. Sch{\"{o}}nberger and Jan{-}Michael Frahm.
\newblock Structure-from-motion revisited.
\newblock In \emph{{IEEE} Conference on Computer Vision and Pattern
  Recognition}, pages 4104--4113, 2016.

\bibitem[Sitzmann et~al.(2020)Sitzmann, Martel, Bergman, Lindell, and
  Wetzstein]{DBLP:conf/nips/SitzmannMBLW20}
Vincent Sitzmann, Julien N.~P. Martel, Alexander~W. Bergman, David~B. Lindell,
  and Gordon Wetzstein.
\newblock Implicit neural representations with periodic activation functions.
\newblock In \emph{Advances in Neural Information Processing Systems}, 2020.

\bibitem[Smart et~al.(2024)Smart, Zheng, Laina, and
  Prisacariu]{DBLP:journals/corr/abs-2408-13912}
Brandon Smart, Chuanxia Zheng, Iro Laina, and Victor~Adrian Prisacariu.
\newblock Splatt3r: Zero-shot gaussian splatting from uncalibrated image pairs.
\newblock \emph{CoRR}, abs/2408.13912, 2024.

\bibitem[Song et~al.(2024)Song, Zheng, Yuan, Gao, Zhao, He, Gu, and
  Zhao]{DBLP:journals/corr/abs-2403-19615}
Xiaowei Song, Jv Zheng, Shiran Yuan, Huan{-}ang Gao, Jingwei Zhao, Xiang He,
  Weihao Gu, and Hao Zhao.
\newblock {SA-GS:} scale-adaptive gaussian splatting for training-free
  anti-aliasing.
\newblock \emph{CoRR}, abs/2403.19615, 2024.

\bibitem[Suhail et~al.(2022)Suhail, Esteves, Sigal, and
  Makadia]{DBLP:conf/cvpr/SuhailESM22}
Mohammed Suhail, Carlos Esteves, Leonid Sigal, and Ameesh Makadia.
\newblock Light field neural rendering.
\newblock In \emph{{IEEE/CVF} Conference on Computer Vision and Pattern
  Recognition}, pages 8259--8269, 2022.

\bibitem[Tancik et~al.(2020)Tancik, Srinivasan, Mildenhall, Fridovich{-}Keil,
  Raghavan, Singhal, Ramamoorthi, Barron, and
  Ng]{DBLP:conf/nips/TancikSMFRSRBN20}
Matthew Tancik, Pratul~P. Srinivasan, Ben Mildenhall, Sara Fridovich{-}Keil,
  Nithin Raghavan, Utkarsh Singhal, Ravi Ramamoorthi, Jonathan~T. Barron, and
  Ren Ng.
\newblock Fourier features let networks learn high frequency functions in low
  dimensional domains.
\newblock In \emph{Advances in Neural Information Processing Systems}, 2020.

\bibitem[Tancik et~al.(2022)Tancik, Casser, Yan, Pradhan, Mildenhall,
  Srinivasan, Barron, and Kretzschmar]{DBLP:conf/cvpr/TancikCYPMSBK22}
Matthew Tancik, Vincent Casser, Xinchen Yan, Sabeek Pradhan, Ben~P. Mildenhall,
  Pratul~P. Srinivasan, Jonathan~T. Barron, and Henrik Kretzschmar.
\newblock Block-nerf: Scalable large scene neural view synthesis.
\newblock In \emph{{IEEE/CVF} Conference on Computer Vision and Pattern
  Recognition}, pages 8238--8248, 2022.

\bibitem[Turki et~al.(2022)Turki, Ramanan, and
  Satyanarayanan]{DBLP:conf/cvpr/TurkiRS22}
Haithem Turki, Deva Ramanan, and Mahadev Satyanarayanan.
\newblock Mega-nerf: Scalable construction of large-scale nerfs for virtual
  fly- throughs.
\newblock In \emph{{IEEE/CVF} Conference on Computer Vision and Pattern
  Recognition}, pages 12912--12921, 2022.

\bibitem[Umeyama(1991)]{DBLP:journals/pami/Umeyama91}
Shinji Umeyama.
\newblock Least-squares estimation of transformation parameters between two
  point patterns.
\newblock \emph{{IEEE} Trans. Pattern Anal. Mach. Intell.}, 13\penalty0
  (4):\penalty0 376--380, 1991.

\bibitem[Ververas et~al.(2024)Ververas, Potamias, Song, Deng, and
  Zafeiriou]{DBLP:journals/corr/abs-2404-19149}
Evangelos Ververas, Rolandos~Alexandros Potamias, Jifei Song, Jiankang Deng,
  and Stefanos Zafeiriou.
\newblock {SAGS:} structure-aware 3d gaussian splatting.
\newblock \emph{CoRR}, abs/2404.19149, 2024.

\bibitem[Wang and Agapito(2024)]{DBLP:journals/corr/abs-2408-16061}
Hengyi Wang and Lourdes Agapito.
\newblock 3d reconstruction with spatial memory.
\newblock \emph{CoRR}, abs/2408.16061, 2024.

\bibitem[Wang et~al.(2021{\natexlab{a}})Wang, Wang, Genova, Srinivasan, Zhou,
  Barron, Martin{-}Brualla, Snavely, and
  Funkhouser]{DBLP:conf/cvpr/WangWGSZBMSF21}
Qianqian Wang, Zhicheng Wang, Kyle Genova, Pratul~P. Srinivasan, Howard Zhou,
  Jonathan~T. Barron, Ricardo Martin{-}Brualla, Noah Snavely, and Thomas~A.
  Funkhouser.
\newblock Ibrnet: Learning multi-view image-based rendering.
\newblock In \emph{{IEEE} Conference on Computer Vision and Pattern
  Recognition}, pages 4690--4699, 2021{\natexlab{a}}.

\bibitem[Wang et~al.(2024)Wang, Leroy, Cabon, Chidlovskii, and
  Revaud]{DBLP:conf/cvpr/Wang0CCR24}
Shuzhe Wang, Vincent Leroy, Yohann Cabon, Boris Chidlovskii, and
  J{\'{e}}r{\^{o}}me Revaud.
\newblock Dust3r: Geometric 3d vision made easy.
\newblock In \emph{{IEEE/CVF} Conference on Computer Vision and Pattern
  Recognition}, pages 20697--20709, 2024.

\bibitem[Wang et~al.(2021{\natexlab{b}})Wang, Wu, Xie, Chen, and
  Prisacariu]{DBLP:journals/corr/abs-2102-07064}
Zirui Wang, Shangzhe Wu, Weidi Xie, Min Chen, and Victor~Adrian Prisacariu.
\newblock Nerf-: Neural radiance fields without known camera parameters.
\newblock \emph{CoRR}, abs/2102.07064, 2021{\natexlab{b}}.

\bibitem[Xia et~al.(2022)Xia, Tang, Timofte, and
  Gool]{DBLP:journals/corr/abs-2210-04553}
Yitong Xia, Hao Tang, Radu Timofte, and Luc~Van Gool.
\newblock Sinerf: Sinusoidal neural radiance fields for joint pose estimation
  and scene reconstruction.
\newblock \emph{CoRR}, abs/2210.04553, 2022.

\bibitem[Xu et~al.(2023)Xu, Xiangli, Peng, Pan, Zhao, Theobalt, Dai, and
  Lin]{DBLP:conf/cvpr/XuXPPZT0L23}
Linning Xu, Yuanbo Xiangli, Sida Peng, Xingang Pan, Nanxuan Zhao, Christian
  Theobalt, Bo Dai, and Dahua Lin.
\newblock Grid-guided neural radiance fields for large urban scenes.
\newblock In \emph{{IEEE/CVF} Conference on Computer Vision and Pattern
  Recognition}, pages 8296--8306, 2023.

\bibitem[Yan et~al.(2024)Yan, Low, Chen, and Lee]{DBLP:conf/cvpr/YanLCL24}
Zhiwen Yan, Weng~Fei Low, Yu Chen, and Gim~Hee Lee.
\newblock Multi-scale 3d gaussian splatting for anti-aliased rendering.
\newblock In \emph{{IEEE/CVF} Conference on Computer Vision and Pattern
  Recognition}, pages 20923--20931, 2024.

\bibitem[Ye et~al.(2024{\natexlab{a}})Ye, Liu, Xu, Xueting, Pollefeys, Yang,
  and Songyou]{ye2024noposplat}
Botao Ye, Sifei Liu, Haofei Xu, Li Xueting, Marc Pollefeys, Ming-Hsuan Yang,
  and Peng Songyou.
\newblock No pose, no problem: Surprisingly simple 3d gaussian splats from
  sparse unposed images.
\newblock \emph{arXiv preprint arXiv:2410.24207}, 2024{\natexlab{a}}.

\bibitem[Ye et~al.(2024{\natexlab{b}})Ye, Li, Kerr, Turkulainen, Yi, Pan,
  Seiskari, Ye, Hu, Tancik, and
  Kanazawa]{ye2024gsplatopensourcelibrarygaussian}
Vickie Ye, Ruilong Li, Justin Kerr, Matias Turkulainen, Brent Yi, Zhuoyang Pan,
  Otto Seiskari, Jianbo Ye, Jeffrey Hu, Matthew Tancik, and Angjoo Kanazawa.
\newblock gsplat: An open-source library for {Gaussian} splatting.
\newblock \emph{arXiv preprint arXiv:2409.06765}, 2024{\natexlab{b}}.

\bibitem[Yu et~al.(2021)Yu, Li, Tancik, Li, Ng, and
  Kanazawa]{DBLP:conf/iccv/YuLT0NK21}
Alex Yu, Ruilong Li, Matthew Tancik, Hao Li, Ren Ng, and Angjoo Kanazawa.
\newblock Plenoctrees for real-time rendering of neural radiance fields.
\newblock In \emph{{IEEE/CVF} International Conference on Computer Vision},
  pages 5732--5741, 2021.

\bibitem[Yu et~al.(2024)Yu, Chen, Huang, Sattler, and
  Geiger]{DBLP:conf/cvpr/YuCHS024}
Zehao Yu, Anpei Chen, Binbin Huang, Torsten Sattler, and Andreas Geiger.
\newblock Mip-splatting: Alias-free 3d gaussian splatting.
\newblock In \emph{{IEEE/CVF} Conference on Computer Vision and Pattern
  Recognition}, pages 19447--19456, 2024.

\end{thebibliography}
}

\appendix
\clearpage
\setcounter{page}{1}
\maketitlesupplementary


\section{Implementation}
\label{supp_sec:implementation}

\paragraph{Neural Scene Regressor.} We initialize part of our model with the pretrained weights from Spann3R~\cite{DBLP:journals/corr/abs-2408-16061}. Following Spann3R, we use a ViT-large encoder, two ViT-base decoders, and a DPT head~\cite{DBLP:conf/iccv/RanftlBK21} for predicting dense pointmaps. We additionally use another DPT head to predict 3D Gaussian primitives. Though Spann3R is trained on images with resolution $224\times 224$, we finetune it on the reconstructed scene with image resolution $512 \times 512$ using AdamW~\cite{DBLP:journals/corr/KingmaB14} optimizer. 

\paragraph{Image Registration.} We use DSAC to register images and compute coarse camera poses. An image is successfully registered if it has at least $5,000$ inliers with a reprojection threshold of $6$ px. We use a number of 64 hypotheses and an inlier alpha of $100$. To accelerate registration, the dense pointmaps for each image are downsampled by $4$. To further speed up the training and reduce memory footprint, we do not use all pointmaps from the newly registered images to refine camera poses. Instead, we pre-build sparse point tracks $\mathbf{X}_k = \{ (\mathbf{u}_{ij}, \mathbf{I}_i) \}$, where $\mathbf{u}_{ij}$ denotes the $j$-th pixels observed on image $\mathbf{I}_i$. We adopt the Union-Find algorithm to remove duplicate and ambiguous tracks to improve the robustness during refinement. We adopt the Huber loss with a threshold of $0.1$ as the robust loss function in Eq.~\eqref{eq:point_to_camera_ray_loss}.

\paragraph{Run Time and Memory Footprint.} Our pipeline converges faster than COLMAP~\cite{DBLP:conf/cvpr/SchonbergerF16}. On the LLFF dataset, our method converges in two epochs, which takes about 25 minutes for each scene; On the MipNeRF360 dataset and the Tanks-and-Temples dataset, our method converges in $5-15$ epochs, which takes about 2 hours for each scene. During training, we evaluate the model and save intermediate results to disks for every $1,000$ iteration. The evaluation time is also included in the training step. Our method takes about 21GB with batch size 1 during training on an NVIDIA 4090 GPU.

\paragraph{Pseudo Algorithm of Our Training Pipeline.} We provide the pseudo algorithm of our incremental training pipeline as described in Sec.~\ref{subsec:incremental_recon} in Alg.~\ref{alg:inc_neural_recon_alg}. At line $1$, $\mathcal{V}_i$ denotes the set of graph nodes, and $\mathcal{E}_i$ denotes the set of graph edges. At line 4, $|\cdot|$ denotes the capacity of a set. We align our predicted camera poses to pseudo-ground-truth using the Umeyama~\cite{DBLP:journals/pami/Umeyama91} algorithm. Note that the camera poses and sparse points of COLMAP are normalized at the end of reconstruction. In line 14, we also normalize our predicted camera poses and dense points before refining the neural scene for fair comparison. We experimentally found this can improve the training stability of 3DGS.

\begin{algorithm}
 \caption{Incremental Neural Reconstruction Algorithm}
 \label{alg:inc_neural_recon_alg}
 
 \begin{algorithmic}[1]
 \Require a set of (unordered) images $\{\mathbf{I}_i\}$, maximum iteration per epoch $\text{iter}_{\text{max}}$
 \Ensure Camera poses $\{\mathbf{T}_k\}$, 3D Gaussian primitives $\{\mathbf{G}_k\}$
 
 \State Construct a similarity graph $\mathcal{G}_{\text{sim}} = (\mathcal{V}_i, \mathcal{E}_i)$
 \State Initialization from a seed image $\textbf{I}_{\text{seed}}$ (\cf~\cref{subsubsec:seed_init})
 \State Registered image set $(\mathcal{I}_{\text{reg}}, \mathcal{T}_{\text{reg}}) = \{(\textbf{I}_{\text{seed}}, \textbf{T}_{\text{seed}})\}$

 \While{$|\mathcal{I}_{\text{reg}}| < |\{\mathbf{I}_i\}|$}
    \State Register a new images $(\mathcal{I}_{\text{new}},\mathcal{T}_{\text{new}})$ by Eq.~\eqref{eq:coarse_camera_pose}
    \State Refine newly registered camera poses by Eq.~\eqref{eq:point_to_camera_ray_loss}
    \State Update training buffer using $(\mathcal{I}_{\text{new}},\mathcal{T}_{\text{new}})$
    \State $\text{iter} := 0$
    \While{$\text{iter} < \text{iter}_{\text{max}}$}
        \State Finetune scene regressor $f_{\text{SCR}}$ using Eq.~\eqref{eq:rgb_loss}
        \State $\text{iter} := \text{iter} + 1$
    \EndWhile
    
    \State $\mathcal{I}_{\text{reg}} := \mathcal{I}_{\text{reg}} + \mathcal{I}_{\text{new}},\ \mathcal{T}_{\text{reg}} := \mathcal{T}_{\text{reg}} + \mathcal{T}_{\text{new}}$
    
 \EndWhile

 \State Finalize camera poses $\mathcal{T}_{\text{reg}} =\{\mathbf{T}_k\}$ using~\eqref{eq:point_to_camera_ray_loss}
 \State Normalize camera poses $\mathcal{T}_{\text{reg}}^{\text{norm}}=\text{Normalize}(\mathcal{T}_{\text{reg}})$
 \State Finalize neural scene $\{\mathbf{G}_k\}$ (\cf~\cref{subsubsec:finalize_neural_scene})

 \end{algorithmic}
\end{algorithm}
\vspace{-3mm}

\section{Additional Results}

\begin{figure*}[htbp]
    \centering

    \includegraphics[width=1\linewidth]{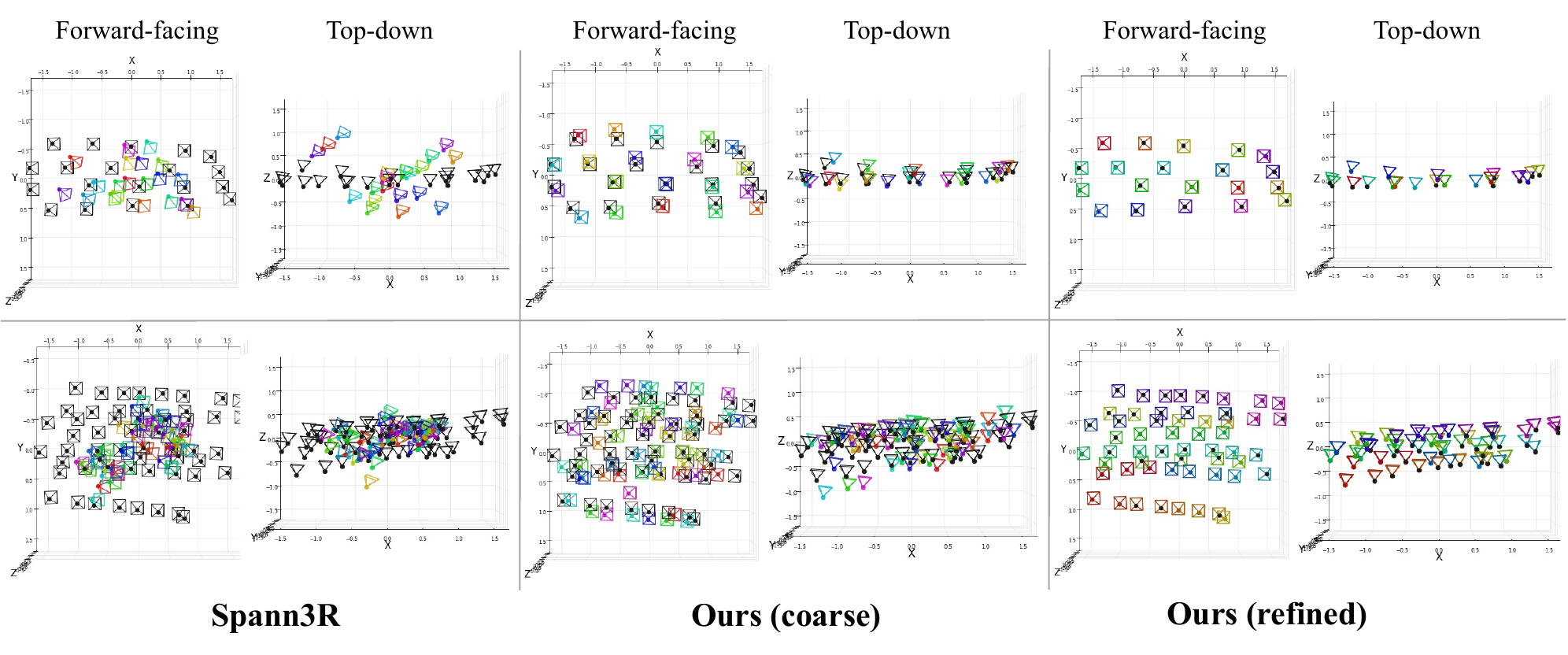}
    \vspace{-7mm}
    \caption{\textbf{Ablation of camera poses refinement on the LLFF dataset} (Zoom in for best view).
            Top and bottom are respectively the camera poses of the `Fern' and `Room' scenes.
     }
    \label{fig:llff_ablation_study}
    \vspace{-3mm}
\end{figure*}

\begin{figure*}[htbp]
    \centering
   
    \includegraphics[width=1.0\linewidth]{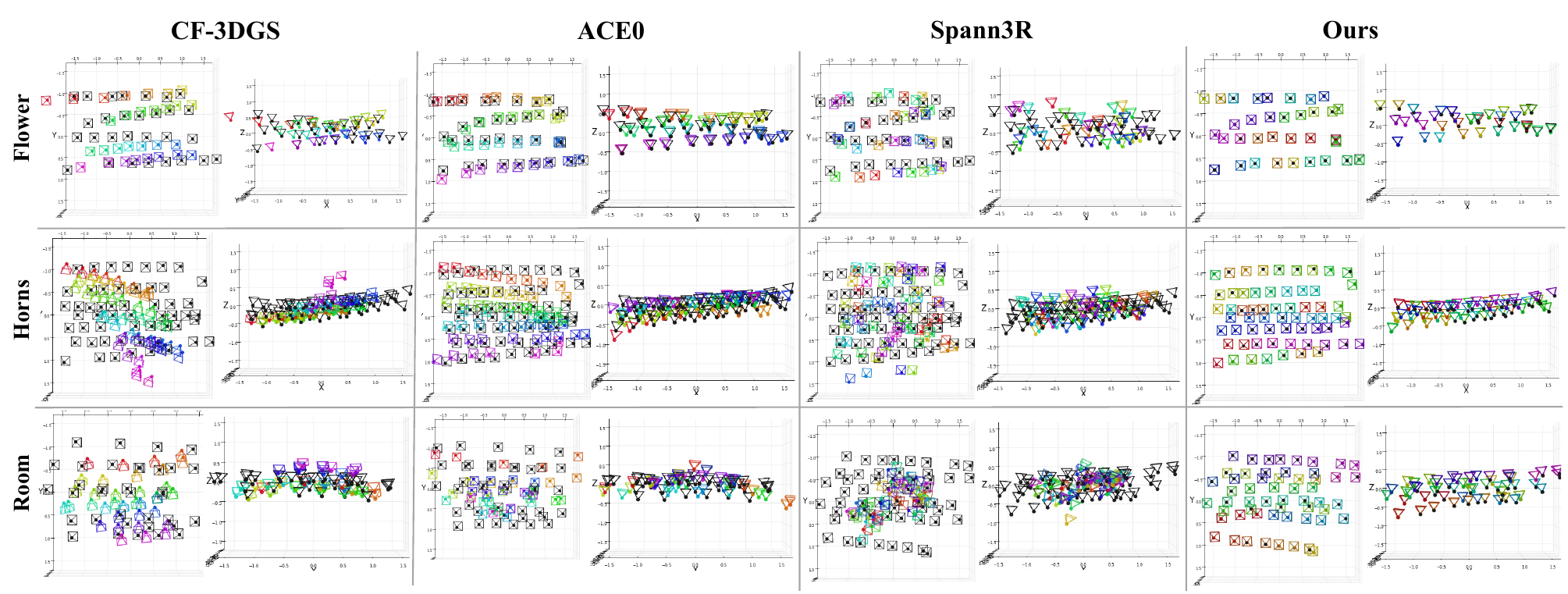}
    \includegraphics[width=1.0\linewidth]{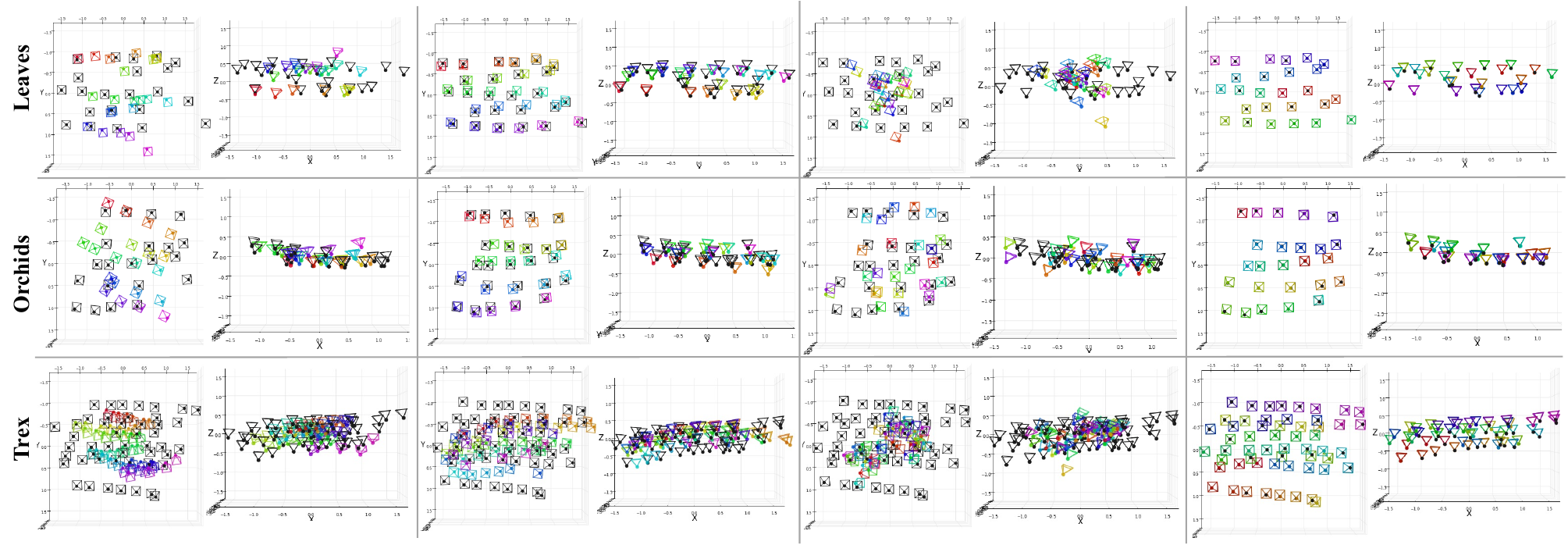}
    \caption{More \textbf{qualitative comparisons of camera poses accuracy on the LLFF dataset} (Zoom in for            best view). Black: pseudo-ground-truth camera poses obtained from COLMAP~\cite{DBLP:conf/cvpr/SchonbergerF16}. Colored: predicted camera poses.}
    \label{fig:llff_pose_cmp_more}
    \vspace{-3mm}
\end{figure*}

\paragraph{Ablation of Pose Refinement.} We present the visual comparison of our method with ($\text{Ours}_{\text{refine}}$) and without ($\text{Ours}_{\text{coarse}}$) the refinement step in Fig.~\ref{fig:llff_ablation_study}. As is shown in Fig.~\ref{fig:llff_ablation_study}, after camera pose refinement, camera poses are aligned closer to ground truth. Compared to the camera poses obtained from Spann3R of the first row in Fig.~\ref{fig:llff_pose_cmp} and the third row in Fig.~\ref{fig:llff_pose_cmp_more}, the coarse camera poses are closer to the ground truth, which is coherent with the quantitative results provided in Table~\ref{table:ablation_study_llff_camera_pose_acc}.

\paragraph{Ablation of Finalizing Reconstruction.}
We ablate the effectiveness of finalizing the camera poses in our training pipeline in Table~\ref{table:quantitative_mipnerf360_camera_pose_finalize_ablation}. The unit for rotation error is degree. We denote our method without the finalizing step as $\text{Ours}_{\text{nf}}$. We can see that the finalization effectively mitigates the error accumulation in both camera rotations and translations. We also emphasize that the camera pose finalizing step is important to the neural scene refinement step since 3DGS is sensitive to even small perturbance camera poses. Moreover, jointly optimizing explicit 3DGS and camera poses during training has limited effect when camera poses are close to ground truth and can even diverge the training~\cite{ye2024gsplatopensourcelibrarygaussian}.

\begin{table}[htbp]
    \centering

    \resizebox{0.5\textwidth}{!}{
        \begin{tabular}{l | c c  c c  c c  c c}
        \toprule

        \multirow{2}{*}{\textbf{Scenes}} &
        \multicolumn{2}{c}{\textbf{Bicycle}} &  
        \multicolumn{2}{c}{\textbf{Counter}} &  
        \multicolumn{2}{c}{\textbf{Garden}}  &  
        \multicolumn{2}{c}{\textbf{Kitchen}} \\ 
        
        \cmidrule(r){2-3} \cmidrule(r){4-5} \cmidrule(r){6-7} \cmidrule(r){8-9}

        & \multirow{1}{*}{$\Delta \mathbf{R}$}
        & \multirow{1}{*}{$\Delta \mathbf{t}$} 
        
        & \multirow{1}{*}{$\Delta \mathbf{R}$}
        & \multirow{1}{*}{$\Delta \mathbf{t}$} 
       
        & \multirow{1}{*}{$\Delta \mathbf{R}$}
        & \multirow{1}{*}{$\Delta \mathbf{t}$} 
    
        & \multirow{1}{*}{$\Delta \mathbf{R}$}
        & \multirow{1}{*}{$\Delta \mathbf{t}$}  \\
       
        \midrule
       
        $\text{Ours}_{\text{nf}}$
        & 0.096 & 0.015                              
        & 0.041 & 0.007
        & 0.095 & 0.012
        & 0.150 & 0.219 \\
        
        Ours 
        & \cellcolor{red}0.035 & \cellcolor{red}0.005  
        & \cellcolor{red}0.029 & \cellcolor{red}0.002  
        & \cellcolor{red}0.028 & \cellcolor{red}0.002  
        & \cellcolor{red}0.052 & \cellcolor{red}0.008  
        \\       
        \bottomrule

        \end{tabular}
    }
    \vspace{-3mm}
    \caption{\textbf{Quantitative results of camera pose accuracy on MipNeRF360 dataset}.
     \colorbox{red}{Red} color denotes the best results. 
    }
    \label{table:quantitative_mipnerf360_camera_pose_finalize_ablation}
    \vspace{-5mm}
\end{table}

\begin{figure*}[htbp]
    \centering
   
    \includegraphics[width=1.0\linewidth]{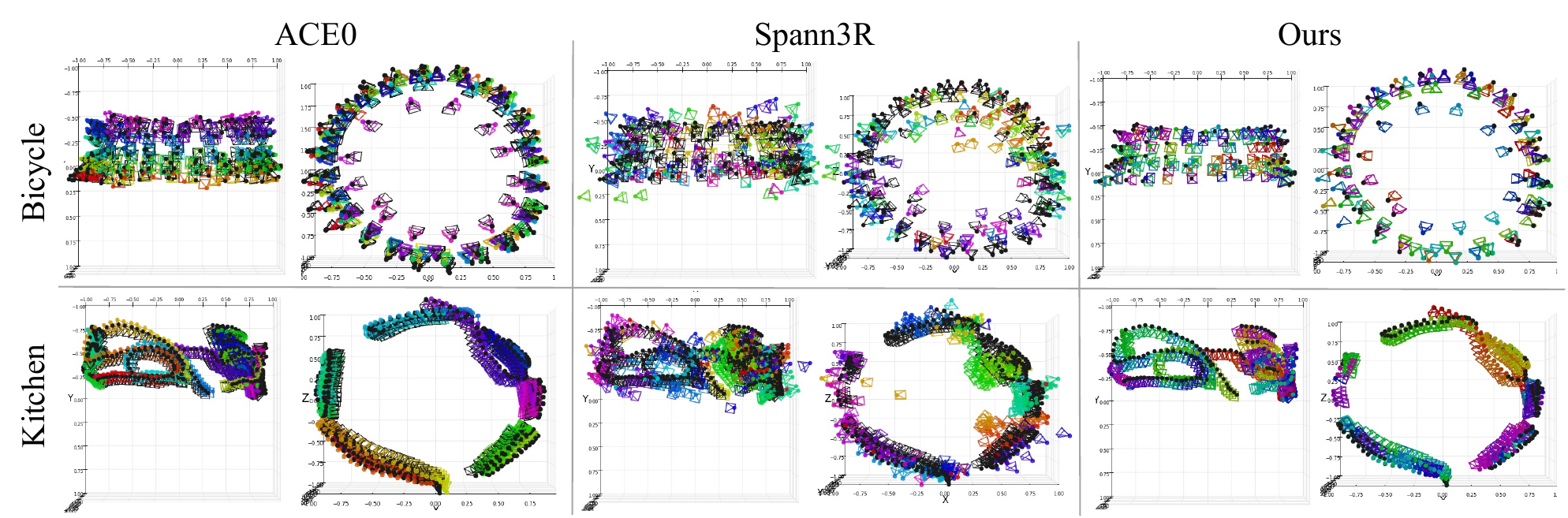}
    \caption{More \textbf{qualitative comparisons of camera poses accuracy on the MipNeRF360 dataset} (Zoom in for best view). Black: pseudo-ground-truth camera poses obtained from COLMAP~\cite{DBLP:conf/cvpr/SchonbergerF16}. Colored: predicted camera poses.
     }
    \label{fig:mipnerf360_pose_cmp_more}
    \vspace{-3mm}
\end{figure*}

\begin{figure*}[htbp]
    \centering
   
    \includegraphics[width=1.0\linewidth]{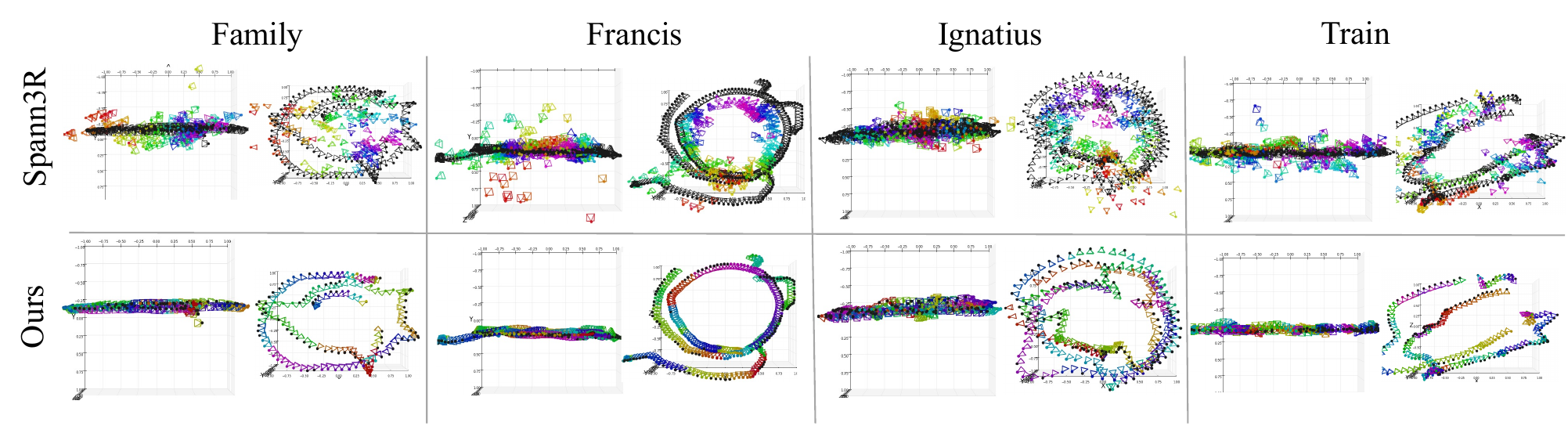}
    \caption{More \textbf{qualitative comparisons of camera poses accuracy on the Tanks-and-Temples dataset} (Zoom in for best view).
     Black: pseudo-ground-truth camera poses obtained from COLMAP~\cite{DBLP:conf/cvpr/SchonbergerF16}. Colored: predicted camera poses.
     }
    \label{fig:tnt_pose_cmp_more}
\end{figure*}

\paragraph{More Results of Camera Poses.} We include the quantitative results of the camera pose accuracy evaluated on the Tanks-and-Temples dataset in Table~\ref{table:quantitative_tnt_camera_pose_acc}. The visual comparison is also provided in Fig.~\ref{fig:tnt_pose_cmp_more}. More visualization results of the aligned camera pose for the LLFF dataset and the MipNeRF360 dataset are respectively provided in Fig.~\ref{fig:llff_pose_cmp_more} and Fig.~\ref{fig:mipnerf360_pose_cmp_more}. We can observe that CF-3DGS~\cite{DBLP:conf/cvpr/Fu0LKKE24} failed to produce faithfully camera poses on the LLFF dataset, which has been analyzed in the main paper. While ACE0~\cite{DBLP:journals/corr/abs-2404-14351} performs very well on the MipNeRF360 dataset, the training is unstable even with a fixed seed number, hence we can not reproduce the comparable result of ACE0 on the Garden scene (\cf Table~\ref{table:quantitative_mipnerf360_camera_pose_acc}) of the MipNeRF360 dataset. Moreover, we find that ACE0 performs poorly on the LLFF dataset. This may be due to the MLP decoder of ACE0 maps individual pixels to 3D space, while it requires well-distributed training views to constrain the network. However, the LLFF dataset contains only forward-facing cameras, which do not provide strong constraints from different view directions and therefore degenerate the training of ACE0.

\begin{table}[htbp]
    \centering

    \resizebox{0.4\textwidth}{!}{
    \begin{tabular}{l | c c  c c}
        \toprule
        
        \multirow{2}{*}{\textbf{Scenes}} &
        \multicolumn{2}{c}{\textbf{Spann3R}~\cite{DBLP:journals/corr/abs-2408-16061}} $\ \ |$ &
        \multicolumn{2}{c}{\textbf{$\text{Ours}$}} \\
        
        \cmidrule(r){2-3} \cmidrule(r){4-5} 
       
        
       
        & \multirow{1}{*}{$\Delta \mathbf{R}$}
        & \multirow{1}{*}{$\Delta \mathbf{t}$} 
    
        & \multirow{1}{*}{$\Delta \mathbf{R}$}
        & \multirow{1}{*}{$\Delta \mathbf{t}$}  \\
       
        \midrule
       
        Family 
        & 16.98 & 0.378 
        & \cellcolor{red}0.036 & \cellcolor{red}0.003 
        \\
        
        Francis 
        & 14.19 & 0.361   
        & \cellcolor{red}0.030 & \cellcolor{red}0.002 
        \\
       
        Ignatius  
        & 11.23 & 0.313  
        & \cellcolor{red}0.028 & \cellcolor{red}0.002  
        \\
    
        Train  
        & 17.33 & 0.286  
        & \cellcolor{red}0.065 & \cellcolor{red}0.011   
        \\

        \bottomrule
    \end{tabular}
    }
    \vspace{-3mm}
    \caption{\textbf{Quantitative results of camera pose accuracy on Tanks-and-Temples dataset}.
     \colorbox{red}{Red} color denotes the best results. 
     }
    \label{table:quantitative_tnt_camera_pose_acc}
    \vspace{-5mm}
\end{table}

\begin{figure*}[htbp]
    \centering
   
    \includegraphics[width=1.0\linewidth]{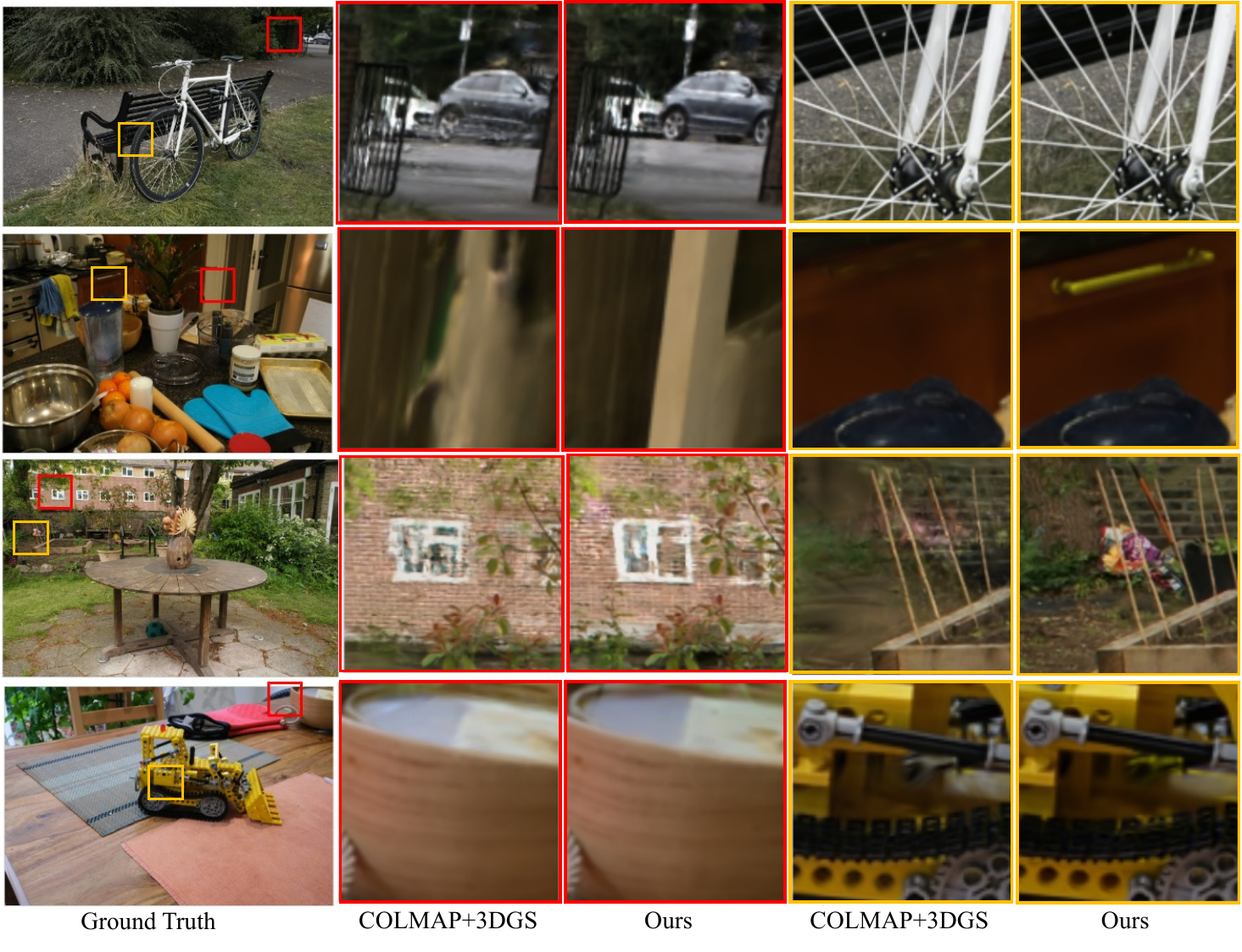}
    \caption{More \textbf{qualitative comparisons of novel view synthesis on the MipNeRF360 dataset}. From top to bottom are respectively the results on scenes of bicycle, counter, garden, and kitchen.}
    \label{fig:mipnerf360_nvs_cmp_more}
\end{figure*}

\begin{figure*}[htbp]
    \centering
   
    \includegraphics[width=1.0\linewidth]{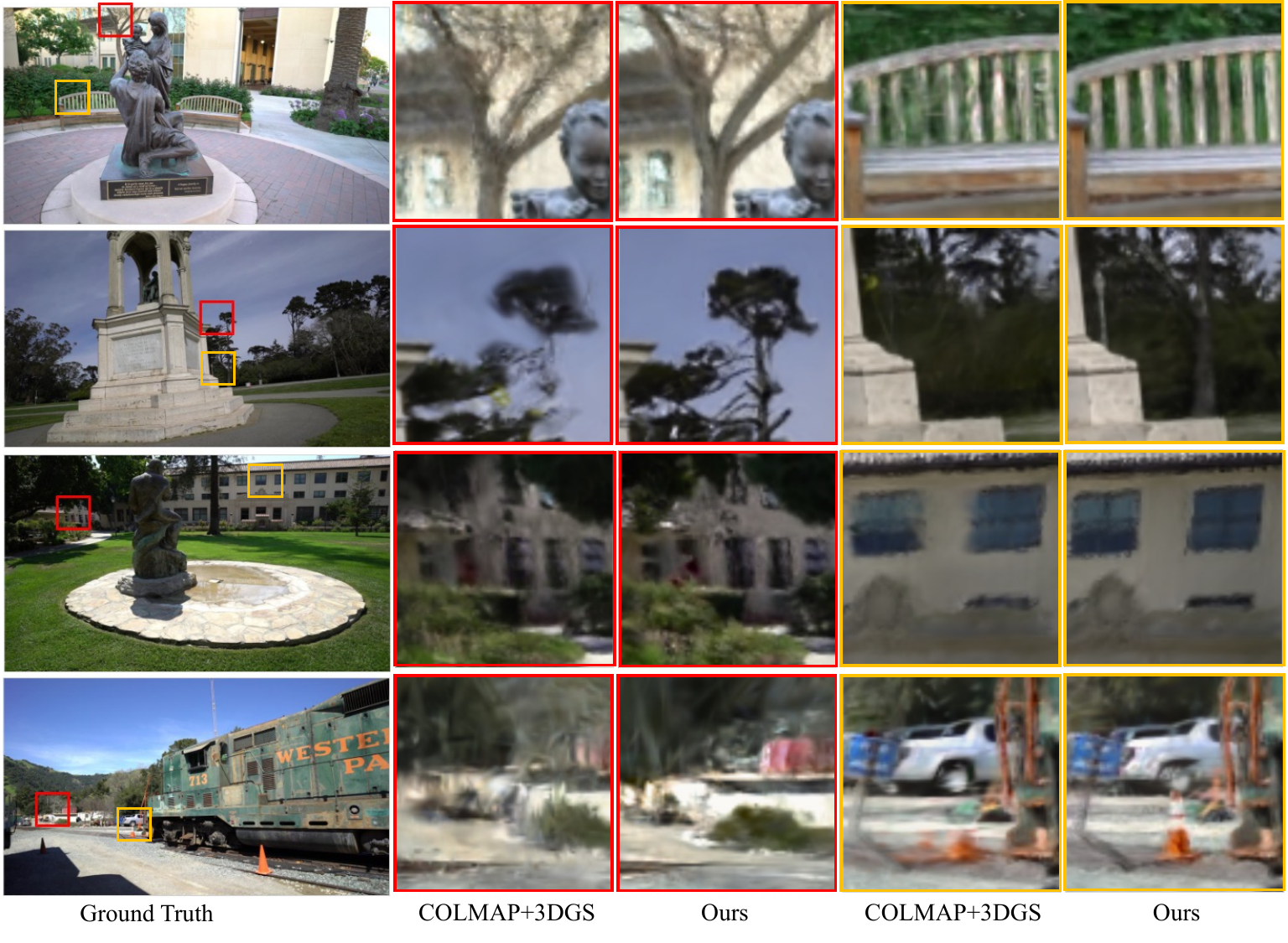}
    \caption{More \textbf{qualitative comparisons of novel view synthesis on the Tanks-and-Temples dataset}. From top to bottom are respectively the results on scenes of the family, Francis, Ignatius, and the train.}
    \label{fig:tnt_nvs_cmp_more}
\end{figure*}

\begin{figure*}[htbp]
    \centering
   
    \includegraphics[width=1.0\linewidth]{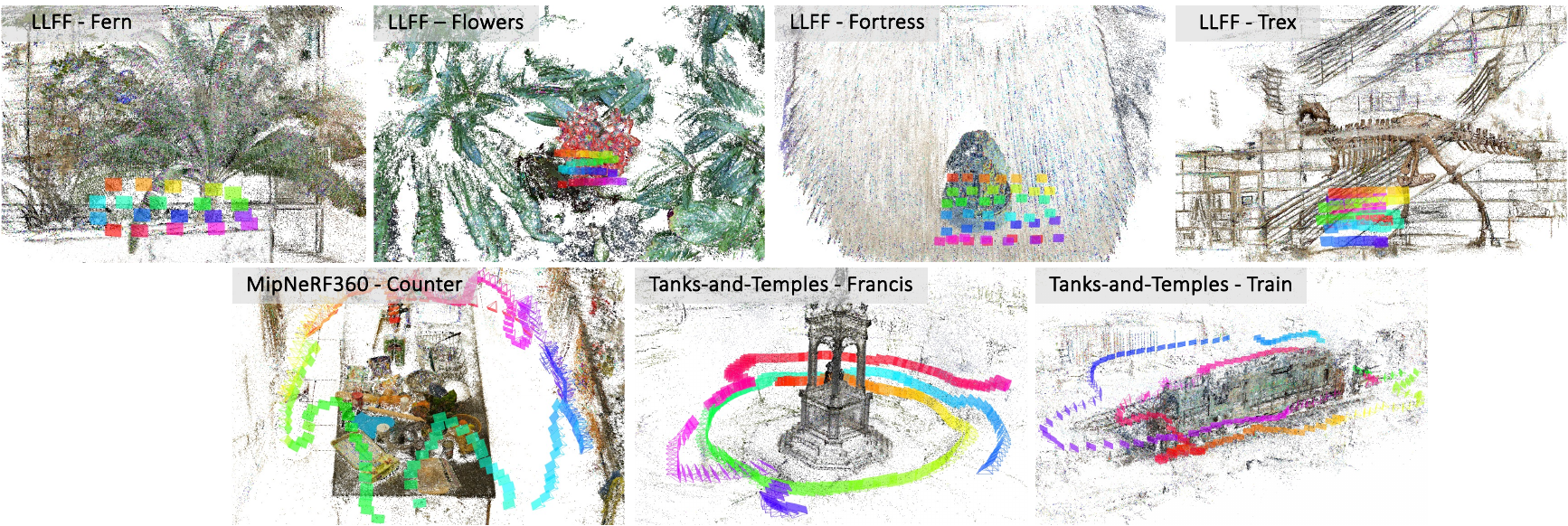}
    \caption{More \textbf{visual reconstruction results} on real-world datasets.}
    \label{fig:more_recon_vis}
\end{figure*}

\paragraph{More Qualitative Results of Novel View Synthesis.} We present more qualitative results of the MipNeRF360 dataset and the Tanks-and-Temples dataset respectively in Fig.~\ref{fig:mipnerf360_nvs_cmp_more} and Fig.~\ref{fig:tnt_nvs_cmp_more} on the novel view synthesis task. The quantitative camera pose accuracy only reveals how close the predicted camera poses to the COLMAP poses. However, it cannot distinguish which one is more accurate since COLMAP poses are only pseudo-ground-truth and it can produce erroneous camera poses. Nonetheless, the results of novel view synthesis provide better metrics to show which one is better when two camera poses are close. In Fig.~\ref{fig:mipnerf360_nvs_cmp_more} and Fig.~\ref{fig:tnt_nvs_cmp_more}, we can observe that our method can render finer details when we zoom into the same areas. The visual comparison also provides coherent support to the quantitative results of novel view synthesis in Table~\ref{table:quantitative_mipnerf360_nvs} and Table~\ref{table:quantitative_tnt_nvs}. More reconstruction results of camera pose and pointmaps are included in Fig.~\ref{fig:more_recon_vis}.

\section{More Discussion}

\paragraph{Limitations.} Though our method can produce higher-quality reconstruction results in terms of both camera poses and novel view synthesis, it requires more GPU memory and training time than ACE0 since we are finetuning transformers, which limits its application on larger scenes. In addition, the training convergence speed of our method relies on the pretrained model of Spann3R. Note that DUSt3R is trained on a mixture of 8 datasets, while Spann3R is only trained on the subset of these datasets.

\paragraph{Future Work.} Our future work includes distilling the pretrained large foundation model into a lightweight network to speed up the training and reduce the GPU memory requirement during training. We will also explore the applicability of the light-weight model on larger-scale and more diverse scenes.

\end{document}